\def\year{2022}\relax
\documentclass[letterpaper]{article} 
\usepackage{aaai22}  
\usepackage{times}  
\usepackage{helvet}  
\usepackage{courier}  
\usepackage[hyphens]{url}  
\usepackage{graphicx} 
\usepackage{natbib}  
\usepackage{caption} 
\DeclareCaptionStyle{ruled}{labelfont=normalfont,labelsep=colon,strut=off} 
\usepackage{algorithm}
\usepackage{algorithmic}

\usepackage{booktabs}
\usepackage{subcaption}
\usepackage{multirow}

\usepackage{newfloat}
\usepackage{listings}
\floatstyle{ruled}
\newfloat{listing}{tb}{lst}{}
\floatname{listing}{Listing}
\title{Landslide Detection in Real-Time Social Media Image Streams}
\author{
    Ferda Ofli,\textsuperscript{\rm 1}
    Muhammad Imran,\textsuperscript{\rm 1}
    Umair Qazi,\textsuperscript{\rm 1}
    Julien Roch,\textsuperscript{\rm 2} \\
    Catherine Pennington,\textsuperscript{\rm 3}
    Vanessa J. Banks,\textsuperscript{\rm 3}
    Remy Bossu \textsuperscript{\rm 2}
    
}
\affiliations{
    \textsuperscript{\rm 1}Qatar Computing Research Institute, \textsuperscript{\rm 2}European-Mediterranean Seismological Centre, \textsuperscript{\rm 3}British Geological Survey\\
    \{fofli, mimran, uqazi\}@hbku.edu.qa, \{julien.roch, bossu\}@emsc-csem.org, \{cpoulton, vbanks\}@bgs.ac.uk
    
}

\begin{document}

\maketitle

\begin{abstract}
Lack of global data inventories obstructs scientific modeling of and response to landslide hazards which are oftentimes deadly and costly. To remedy this limitation, new approaches suggest solutions based on citizen science that requires active participation. However, as a non-traditional data source, social media has been increasingly used in many disaster response and management studies in recent years. Inspired by this trend, we propose to capitalize on social media data to mine landslide-related information automatically with the help of artificial intelligence (AI) techniques. Specifically, we develop a state-of-the-art computer vision model to detect landslides in social media image streams in real time. To that end, we create a large landslide image dataset labeled by experts and conduct extensive model training experiments. The experimental results indicate that the proposed model can be deployed in an online fashion to support global landslide susceptibility maps and emergency response.
\end{abstract}

\section{Introduction}
Landslides occur all around the world and cause thousands of deaths and billions of dollars in infrastructural damage worldwide every year~\cite{kjekstad2009economic}. However, landslide events are often under-reported and insufficiently documented due to their complex natural phenomena governed by various intrinsic and external conditioning and triggering factors such as earthquakes and tropical storms, which are usually more conspicuous, and hence, more widely reported~\cite{lee2004landslide}. Due to this oversight and lack of global data inventories to study landslides, any attempt to quantify global landslide hazards and the associated impacts is destined to be an underestimation~\cite{froude2018global}.

In an attempt to tackle the challenge of building a global landslide inventory, NASA launched a website\footnote{\url{https://gpm.nasa.gov/landslides/index.html}} in 2018 to allow citizens to report about the regional landslides they see in-person or online~\cite{juang2019using}. Following the same idea, researchers further developed other means such as mobile apps to collect citizen-provided data~\cite{kocaman2019citsci,cieslik2019building}. These efforts also help address concerns about news media sources' reporting biases~\cite{moeller2006regarding,pennington2012landslide}. However, this means the bulk of data collection and interpretation still involves time consuming work by specialists searching the Internet for news and social media reports, or directly engaging in communications with those submitting information and then interpreting the data received~\cite{kocaman2019citsci,juang2019using,pennington2015national,taylor2015enriching}.

To alleviate the need for opt-in participation and manual processing, we strive to develop a state-of-the-art AI model that can automatically detect landslides\footnote{By landslides, we refer to all downward and outward movement of loosen slope materials such as landslip, debris flows, mudslides, rockfalls, earthflows, and other mass movements.} in social media image streams in real time. To achieve this goal, we first create a large image dataset comprising more than 11,000 images from various data sources annotated by domain experts. We then exploit this dataset in a comprehensive experimentation searching for the optimal landslide model configuration. This exploration reveals interesting insights about the model training process. More importantly, the experimental results show that the optimal landslide model achieves a promising performance on a held-out test set. Based on this model, we envision a system that can contribute to harvesting of global landslide data, and hence, facilitate further landslide research. Furthermore, it can support global landslide susceptibility maps to provide situational awareness and improve emergency response and decision making.
 
\section{Related Work}
The literature on landslide detection and mapping approaches mainly uses four types of data sources: \textit{(i) physical sensors}, \textit{(ii) remote sensing}, \textit{(iii) volunteers}, and \textit{(iv) social networks}. Sensor-based approaches rely on land characteristics such as rainfall, altitude, soil type, and slope, to detect landslides and develop models to predict future events~\cite{merghadi2020machine,ramesh2009wireless}. While these approaches can be highly accurate for sub-catchments to the referenced data area, their large-scale deployment is extremely costly.

Earth observation data obtained using high-resolution satellite imagery has been widely used for landslide detection, mapping, and monitoring~\cite{tofani2013use}. Remote sensing techniques either use Synthetic Aperture Radar (SAR) or optical imagery to perform landslide detection as an image classification, segmentation, object detection, or change detection task~\cite{mohan2021review, cheng2013automatic}. While remote sensing through satellites can be useful to monitor landslides globally, their deployment can prove costly and time-consuming. Moreover, satellite data is susceptible to noise such as clouds.

A few studies demonstrate the use of Volunteered Geographical Information (VGI) as an alternative method to detect landslides~\cite{kocaman2019citsci,can2019convolutional,can2020development}. These studies assume active participation of volunteers to collect landslide data where the volunteers opt in to use a mobile app to provide information such as photos, time of occurrence, damage description and other observations about a landslide event. On the contrary, our work aims to capitalize on massive social media data without any active participation requirement and with better scalability. In addition, we construct a much larger dataset to train deep learning models and perform more extensive experimental evaluations. 

The use of social media data for landslide detection has not been explored extensively. To the best of our knowledge, no prior work has explored the use of social media imagery to detect landslides. The most relevant work reported in~\cite{musaev2014litmus, musaev2017rex} combines social media text data and physical sensors to detect landslides. The authors used textual messages collected through a set of landslide-related keywords on Twitter, Instagram, and YouTube, which were then combined with sensor data about seismic activity and rainfall to train a machine learning classifier that can identify landslide incidents. In this study, we focus on analyzing social media images which can provide more detailed information about the impact of the landslide event. To that end, our work can be considered as complementary to prior art.





\section{Dataset}

\begin{figure*}
	\renewcommand{\arraystretch}{0.6} 
	\linespread{0.5}\selectfont\centering
	\resizebox{0.99\linewidth}{!}{%
		\begin{tabular}{p{0.17\textwidth} p{0.17\textwidth} p{0.17\textwidth} p{0.17\textwidth} p{0.17\textwidth} p{0.17\textwidth}}
		    \multicolumn{3}{c}{\textbf{LANDSLIDE}} & \multicolumn{3}{c}{\textbf{NOT-LANDSLIDE}}\\
		    \cmidrule(lr){1-3}\cmidrule(lr){4-6}
			\includegraphics[width=0.17\textwidth]{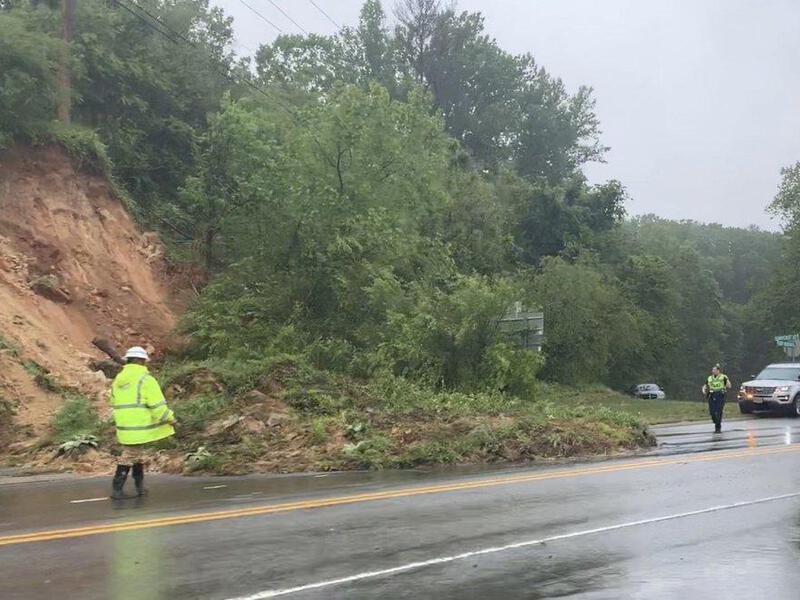}
			&
			\includegraphics[width=0.17\textwidth]{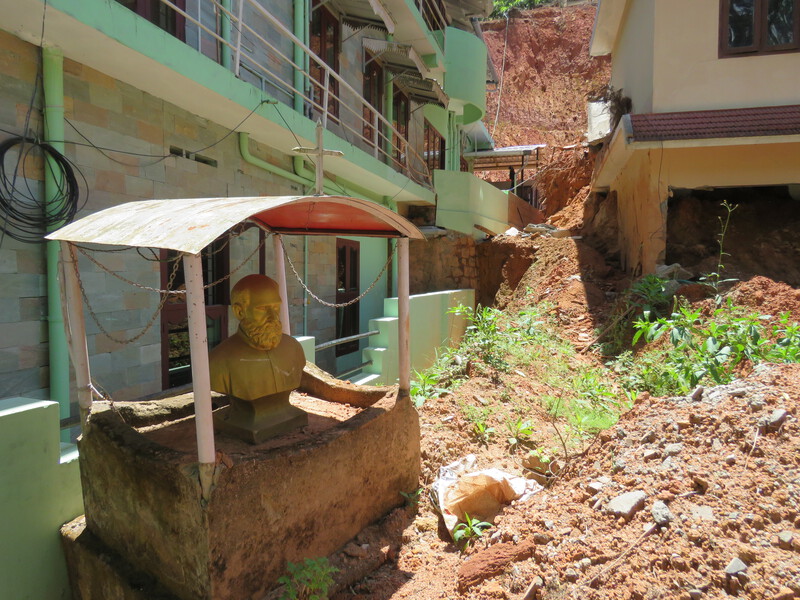}
			&
			\includegraphics[width=0.17\textwidth]{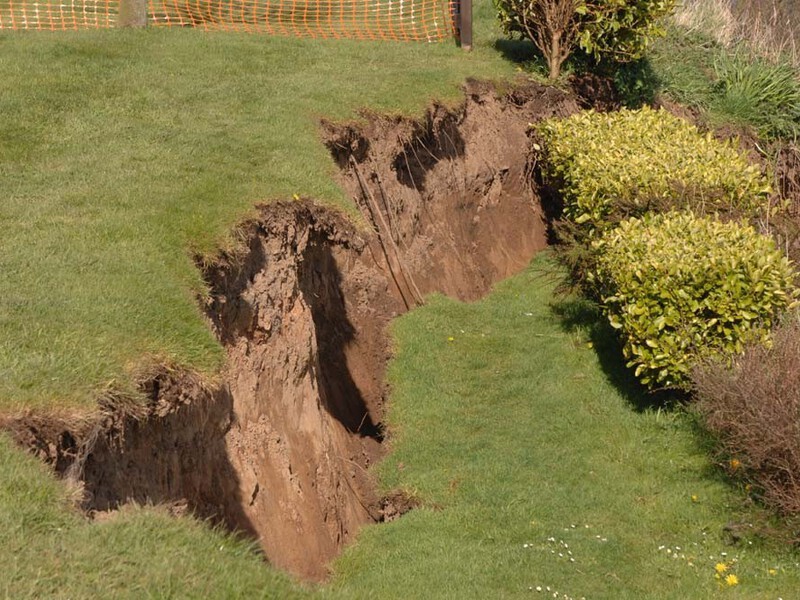}
			&
			\includegraphics[width=0.17\textwidth]{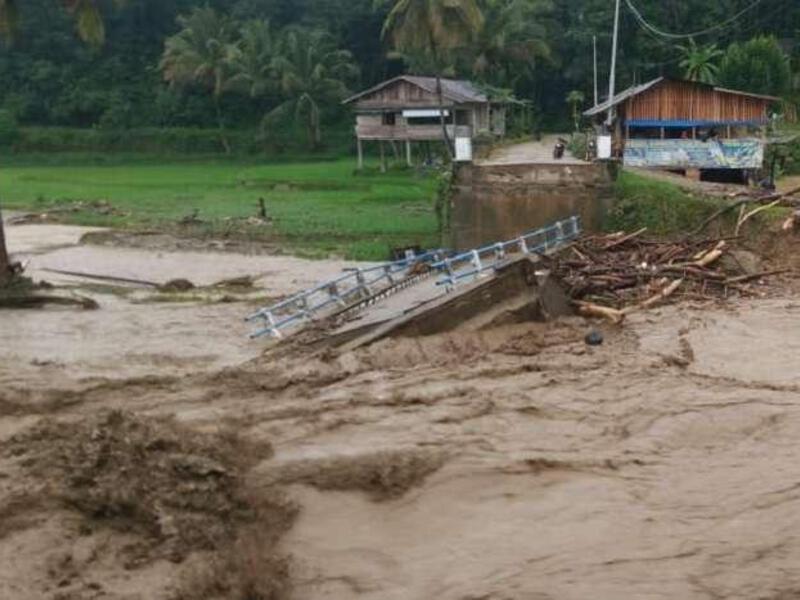}
			&
			\includegraphics[width=0.17\textwidth]{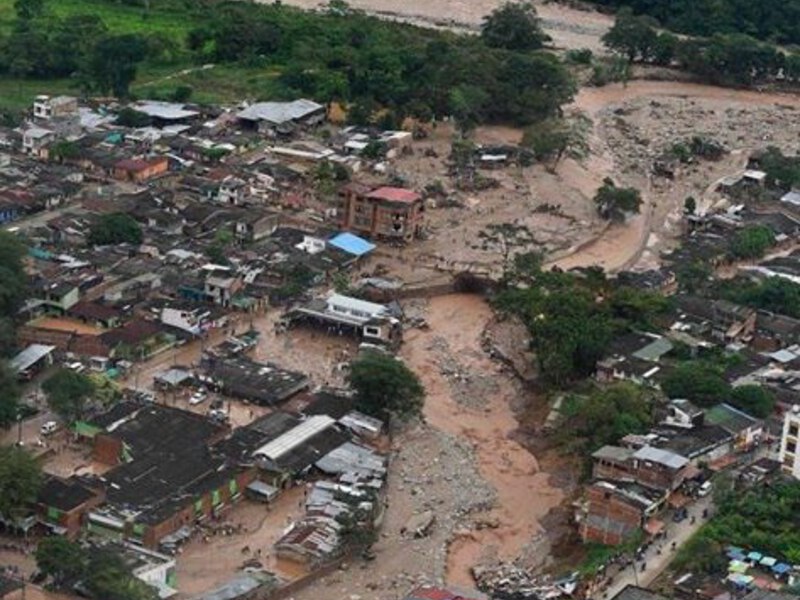}
			&
			\includegraphics[width=0.17\textwidth]{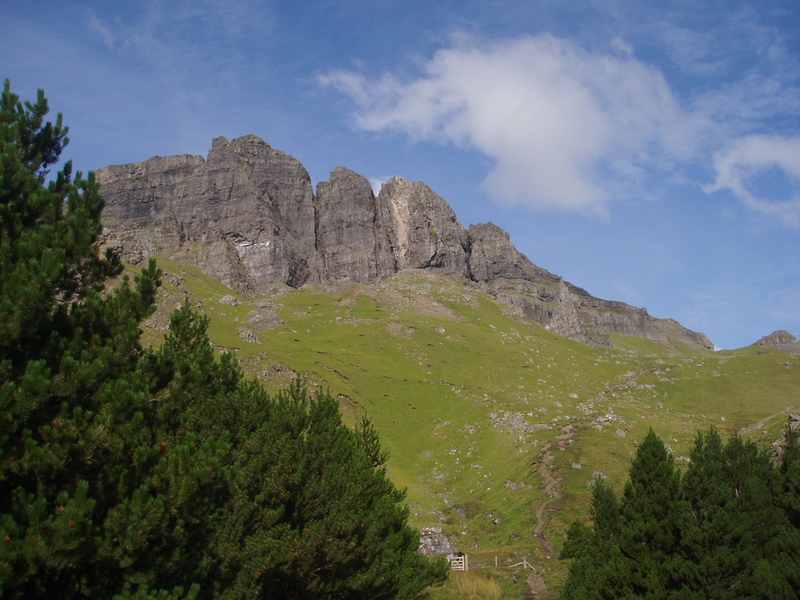}
			\\
			\includegraphics[width=0.17\textwidth]{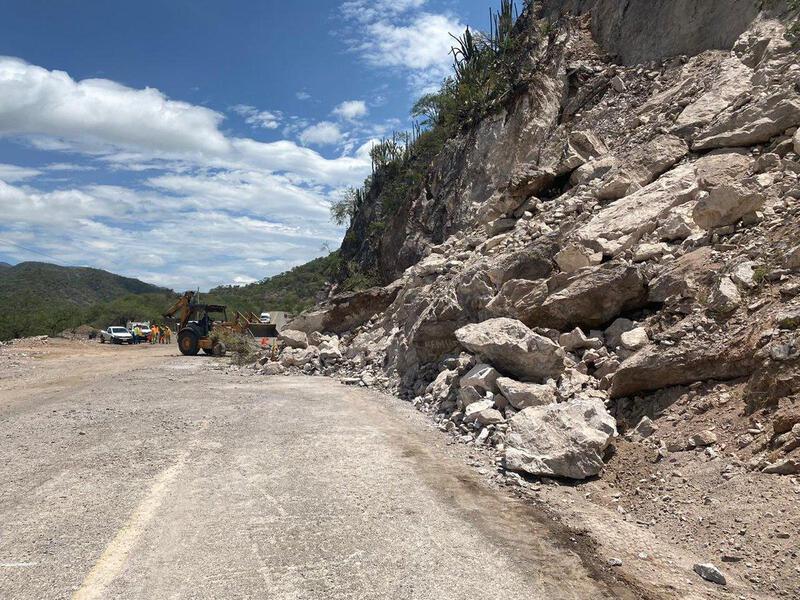}
			&
			\includegraphics[width=0.17\textwidth]{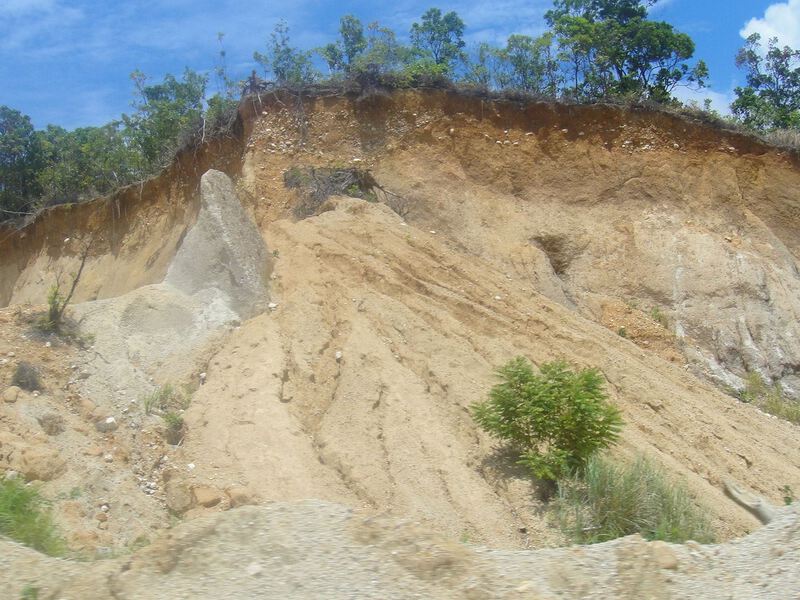}
			&
			\includegraphics[width=0.17\textwidth]{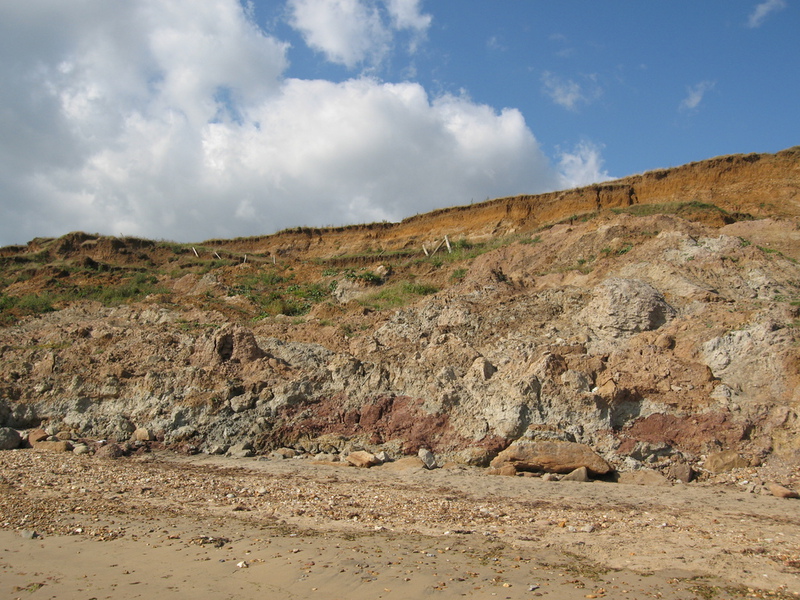}
			&
			\includegraphics[width=0.17\textwidth]{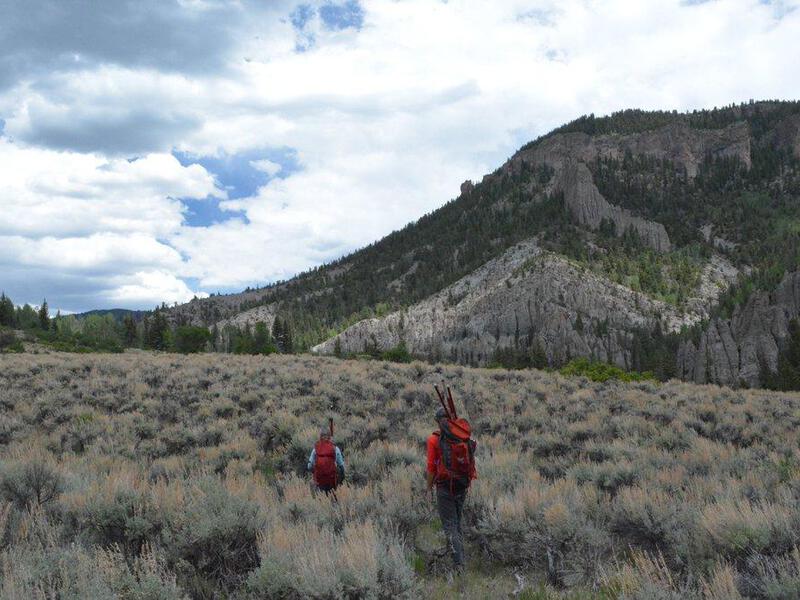}
			&
			\includegraphics[width=0.17\textwidth]{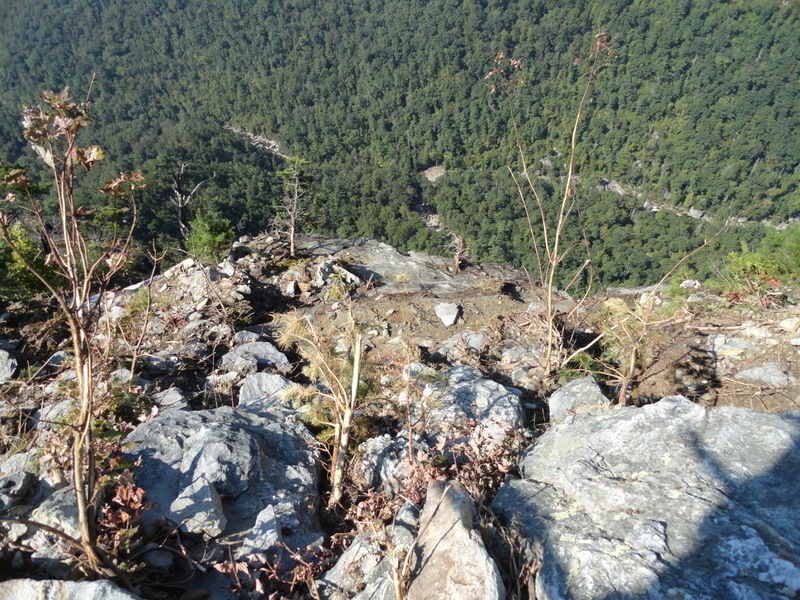}
			&
			\includegraphics[width=0.17\textwidth]{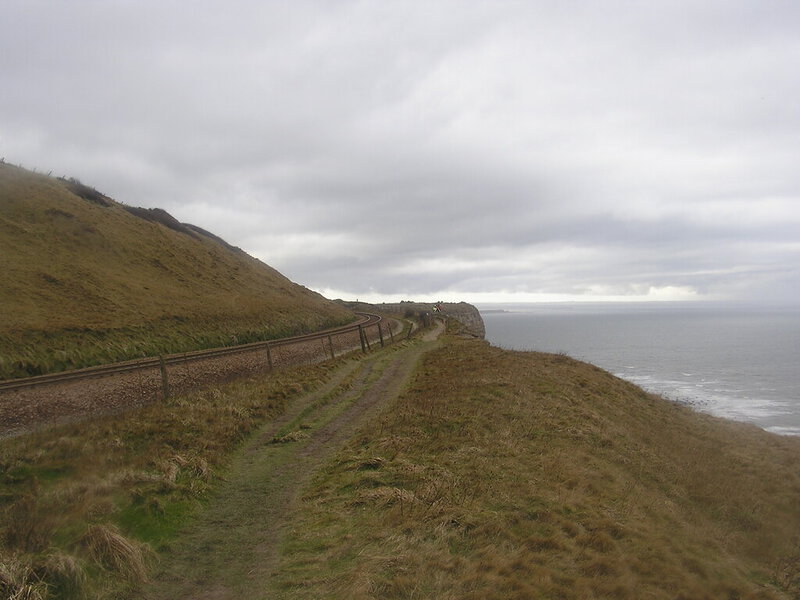}
			\\
			\includegraphics[width=0.17\textwidth]{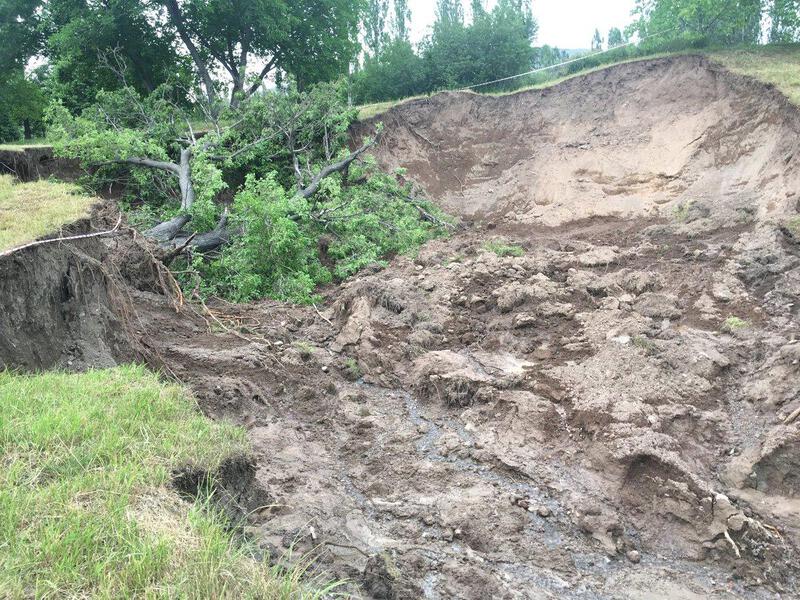}
			&
			\includegraphics[width=0.17\textwidth]{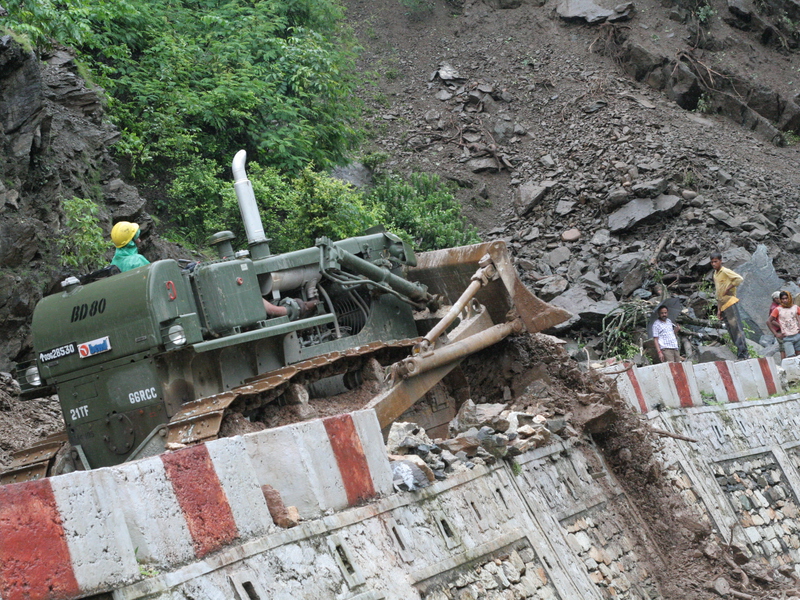}
			&
			\includegraphics[width=0.17\textwidth]{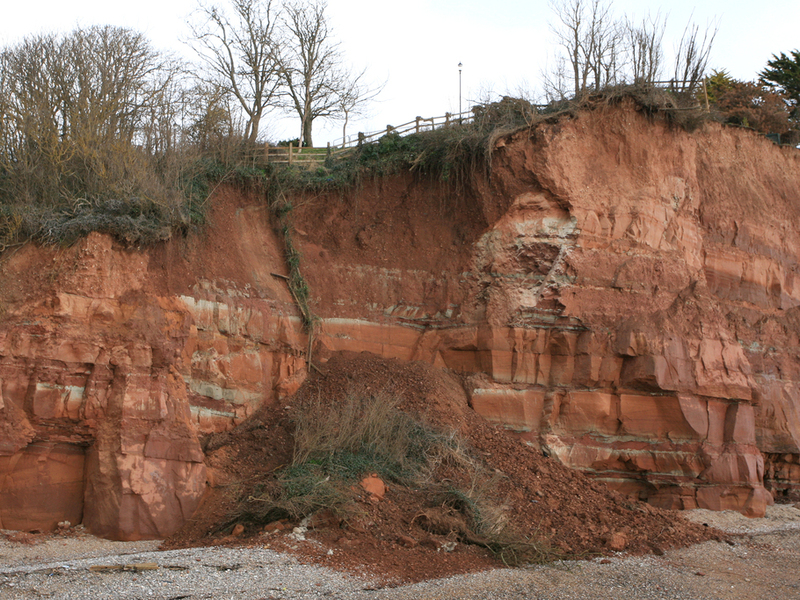}
			&
			\includegraphics[width=0.17\textwidth]{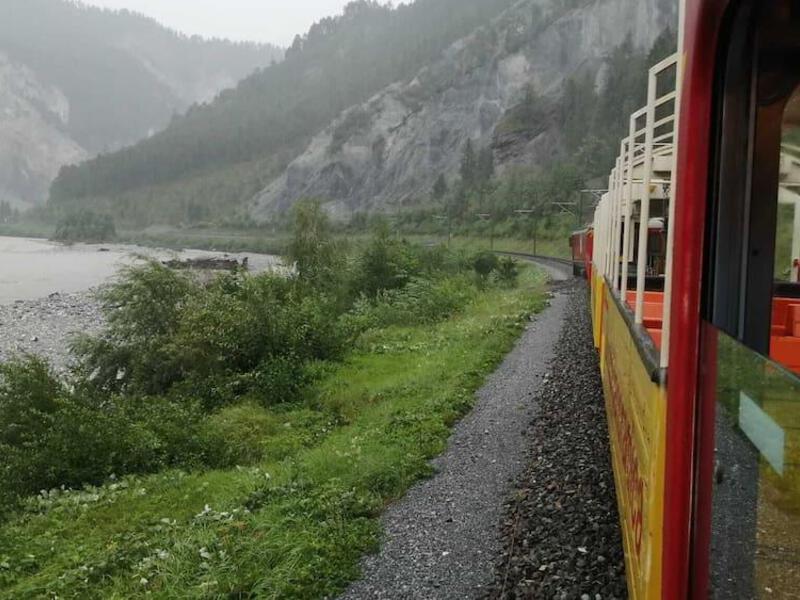}
			&
			\includegraphics[width=0.17\textwidth]{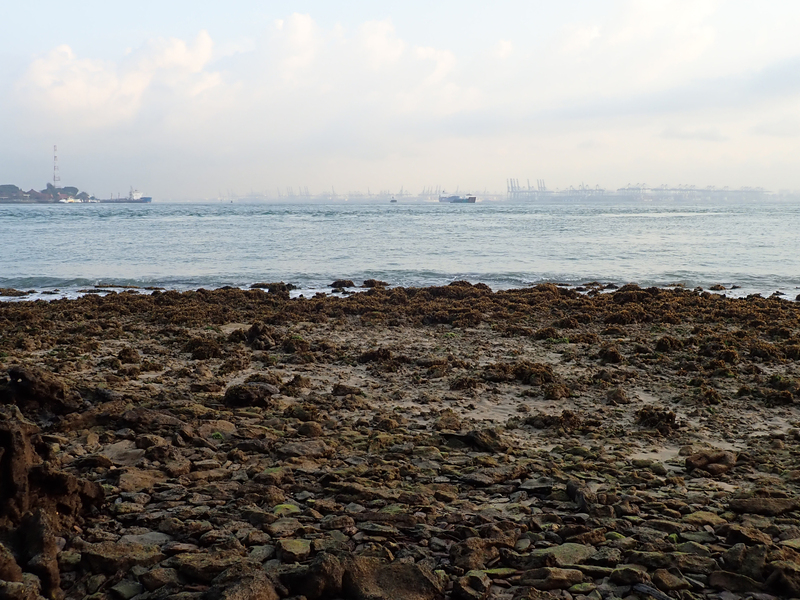}
			&
			\includegraphics[width=0.17\textwidth]{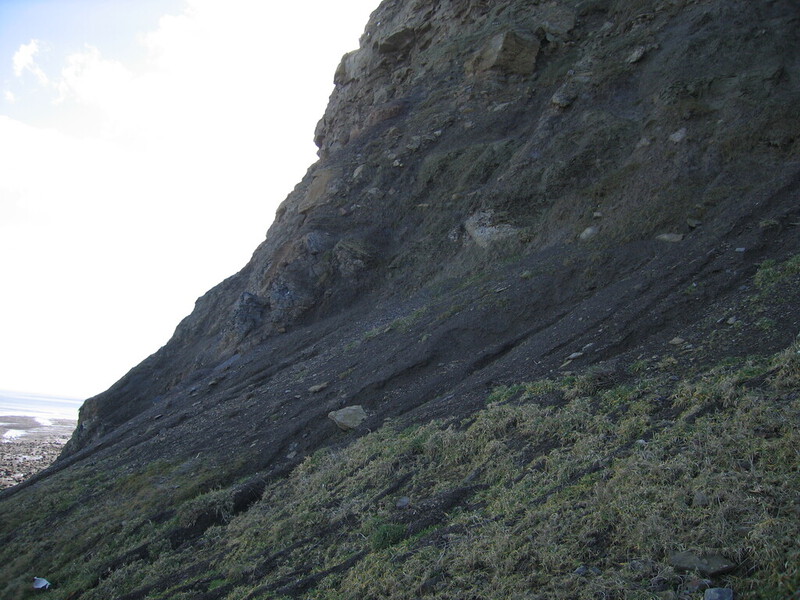}
			\\
			\includegraphics[width=0.17\textwidth]{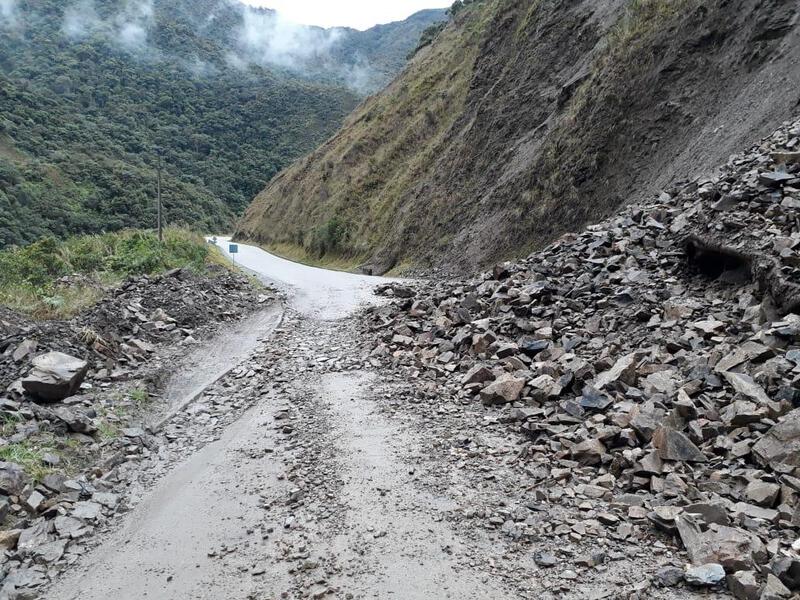}
			&
			\includegraphics[width=0.17\textwidth]{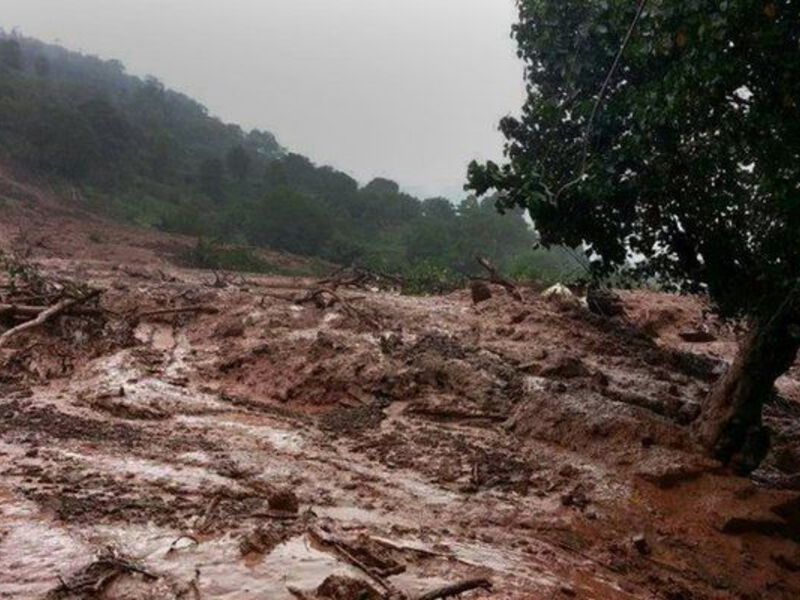}
			&
			\includegraphics[width=0.17\textwidth]{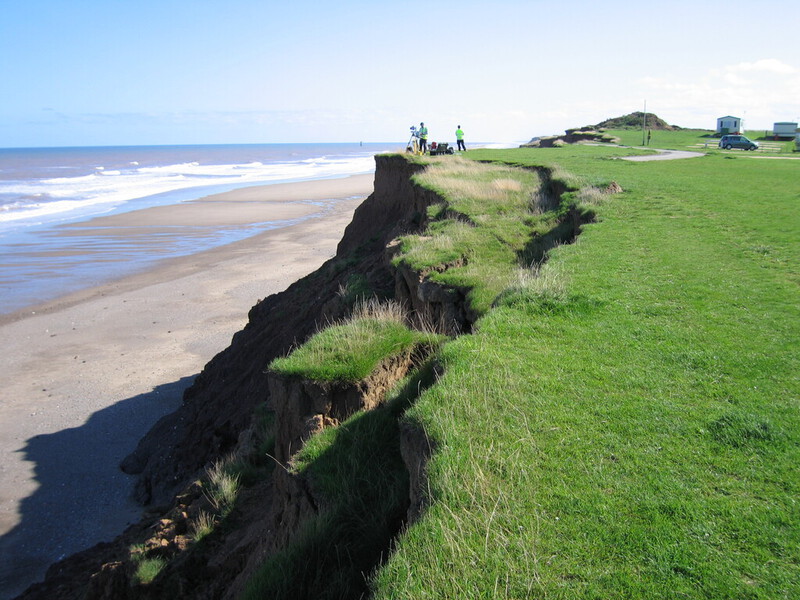}
			&
			\includegraphics[width=0.17\textwidth]{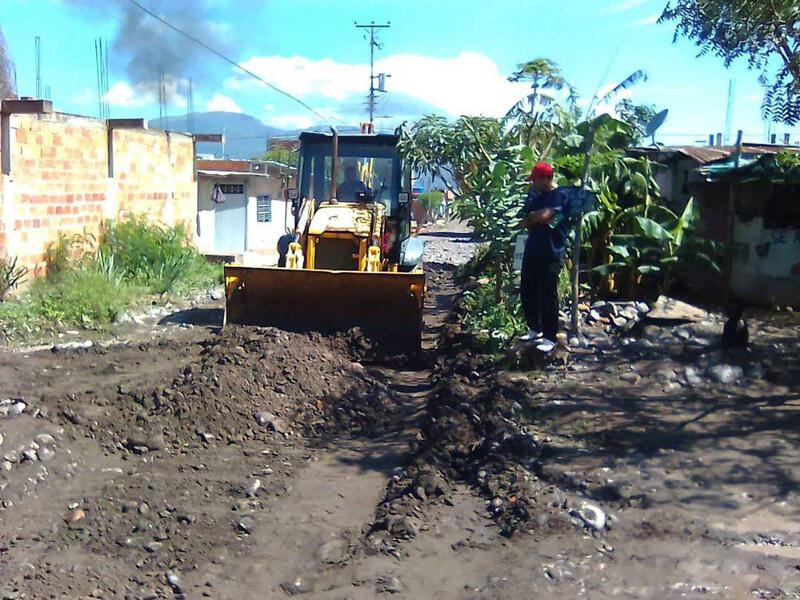}
			&
			\includegraphics[width=0.17\textwidth]{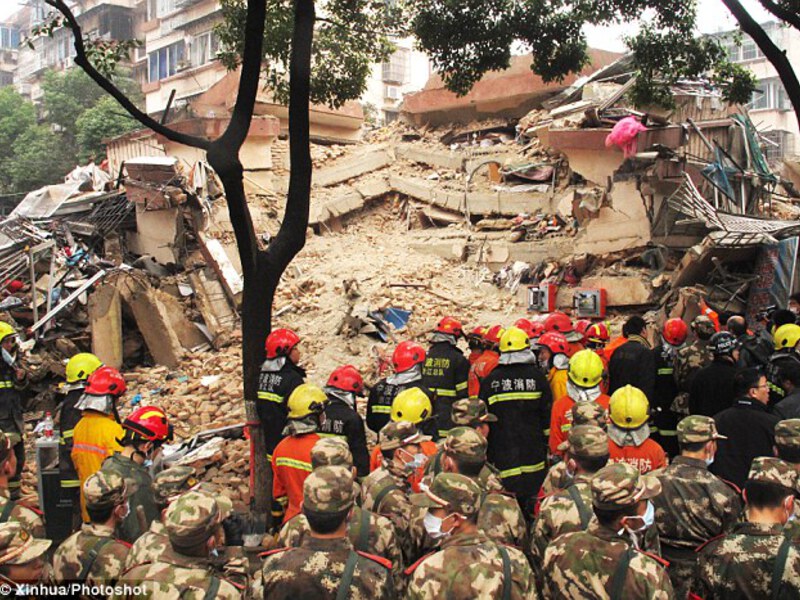}
			&
			\includegraphics[width=0.17\textwidth]{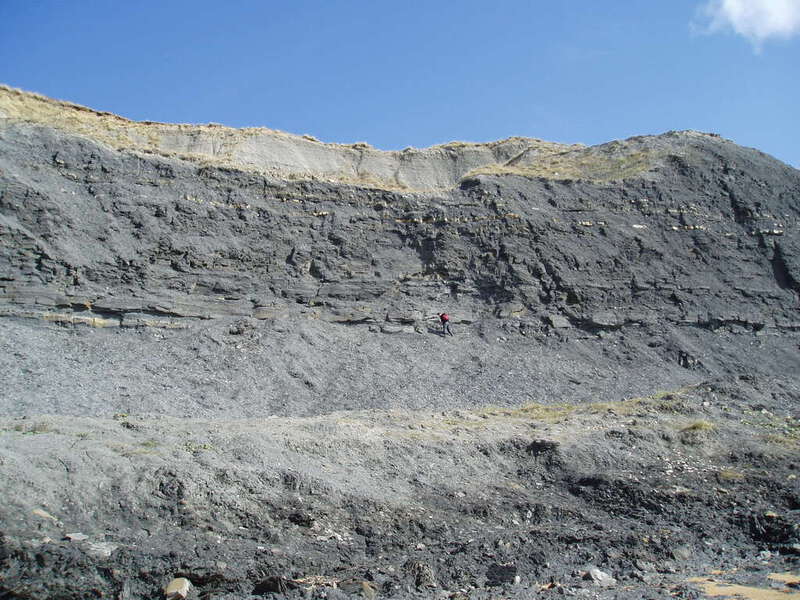}
			\\
			\multicolumn{1}{c}{{\scriptsize \textbf{Twitter}}}
			&
			\multicolumn{1}{c}{{\scriptsize \textbf{Google}}}
			&
			\multicolumn{1}{c}{{\scriptsize \textbf{BGS}}}
			&
			\multicolumn{1}{c}{{\scriptsize \textbf{Twitter}}}
			&
			\multicolumn{1}{c}{{\scriptsize \textbf{Google}}}
			&
			\multicolumn{1}{c}{{\scriptsize \textbf{BGS}}}
			\\
		\end{tabular}
	}%
	\caption{Example images from the dataset.}
	\label{fig:example_images}
\end{figure*}

To train models that can detect landslides in images, we curated a large image dataset from multiple sources with different characteristics. Some images were obtained from the Web using Google Image search with keywords such as \emph{landslide, landslip, earth slip, mudslide, rockslide, rock fall} whereas some images were collected from Twitter using similar landslide-related hashtags. Additional images were obtained from the British Geological Survey archives. The images obtained from social media or the Web are usually noisy and can include duplicates. Therefore, the collected data is manually labeled by three landslide experts who are also co-authors of this study. Since the AI task at hand is ``given an image, recognize landslides'' (i.e., no other external information or expert knowledge is available to the AI model), the experts were instructed to keep this \emph{computer-vision perspective} in mind and label only the most evident cases as ``landslide'' images (i.e., the images where the landslide is the main theme exhibiting substantial visual cues for the computer vision model to learn from). In this context, the BGS images were also included in the labeling process to maintain label consistency across the dataset. On the other hand, since our ultimate goal is to develop a system that will continuously monitor the noisy social media streams to detect landslide events in real time, we retained \emph{negative} (i.e., not-landslide) images that illustrate completely irrelevant cases (e.g., cartoons, advertisements, selfies) as well as difficult scenarios such as post-disaster images from earthquakes and floods in addition to other natural scenes without landslides in the final dataset. Despite the inherent difficulty of the task, the experts achieved an overall Fleiss' Kappa score of 0.58~\cite{fleiss1971measuring}, which indicates an almost substantial inter-annotator agreement. The final dataset contains 11,737 images. Some example images are shown in Figure~\ref{fig:example_images}. The distribution of images across data sources is summarized in Table~\ref{tab:data_sources} and their breakdown into data splits are presented in Table~\ref{tab:data_splits}. As suggested by Table~\ref{tab:data_splits}, only about 23\% of the images are categorized as ``landslide''.

\begin{table}
\centering
\begin{tabular}{@{}lrrrr@{}}
\toprule
 & Training & Validation & Test & Total \\ \midrule
Google & 4,398 & 628 & 1,258 & 6,284 \\
Twitter & 807 & 115 & 231 & 1,153 \\
BGS & 3,010 & 430 & 860 & 4,300 \\ \midrule
Total & 8,215 & 1,173 & 2,349 & 11,737 \\
\bottomrule
\end{tabular}
\caption{Distribution of images across data sources.}
\label{tab:data_sources}
\end{table}

\begin{table}
\centering
\begin{tabular}{@{}lrrrr@{}}
\toprule
 & Training & Validation & Test & Total \\ \midrule
Landslide & 1,883 & 271 & 536 & 2,690 \\
Not-landslide & 6,332 & 902 & 1,813 & 9,047 \\ \midrule
Total & 8,215 & 1,173 & 2,349 & 11,737 \\
\bottomrule
\end{tabular}
\caption{Data splits (70:10:20).}
\label{tab:data_splits}
\end{table}

\section{Landslide Model}

Many computer vision tasks have greatly benefited from the recent advances in deep learning. The features learned in deep convolutional neural networks (CNNs) are proven to be transferable and quite effective when used in other visual recognition tasks~\cite{JDonahue:ICML14, Sermanet:ICLR14, Zeiler:ECCV14, RGirshick:CVPR14, MOquab:CVPR14}, particularly when training samples are limited and learning a successful deep model is not feasible. Considering we also have limited training examples for data-hungry deep CNNs, we follow a transfer learning approach to adapt the features and parameters of the network from the broad domain (i.e., large-scale image classification) to the specific one (i.e., landslide classification). To that end, we conducted extensive experiments where we trained several different deep CNN architectures using different optimizers, learning rates, weight decays, and class balancing strategies.

\smallskip\noindent\textbf{CNN Architecture.} The type of CNN architecture (arch) plays a significant role on the performance of the resulting model depending on the available data size and problem characteristics. Therefore, we explored a representative sample of well-known CNN architectures in our experiments including VGG16~\cite{simonyan2014very}, ResNet18, ResNet50, ResNet101~\cite{he2016deep}, DenseNet~\cite{huang2017densely}, InceptionNet~\cite{szegedy2016rethinking}, and EfficientNet~\cite{tan2019EfficientNet}, among others.

\smallskip\noindent\textbf{Optimizer.} An optimizer (opt) is an algorithm or method that changes the attributes of a neural network (e.g., weights and learning rate) in order to reduce the optimization loss and to increase the desired performance metric (e.g., accuracy). In this study, we experimented with the most popular optimizers, i.e., Stochastic Gradient Descent (SGD) and Adam~\cite{kingma2014adam} with decoupled weight decay regularization~\cite{loshchilov2017decoupled}.

\smallskip\noindent\textbf{Learning rate.} Learning rate (lr) controls how quickly the model is adapted to the problem. Using a too large learning rate can cause the model to converge too quickly to a suboptimal solution whereas a too small learning rate can cause the process to get stuck. Since learning rate is one of the most important hyperparameters and setting it correctly is critical for real-world applications, we performed a grid search over a range of values typically covered in the literature (i.e., $\{10^{-2},10^{-3},10^{-4},10^{-5},10^{-6}\}$).

\smallskip\noindent\textbf{Weight decay.} Weight decay (wd) controls the regularization of the model weights, which in turn, helps to avoid overfitting of a deep neural network on the training data and improve the performance of the model on the unseen data (i.e., better generalization ability). In light of this, we experimented with a range of weight decay values (i.e., $\{10^{-2},10^{-3},10^{-4},10^{-5}\}$).

\smallskip\noindent\textbf{Class balancing.} An imbalanced dataset can bias the prediction model towards the dominant class (i.e., not-landslide) and lead to poor performance on the minority class (i.e., landslide), which would not be ideal for our application. There are many approaches to tackle this problem, ranging from generating synthetic data to using specialized algorithms and loss functions. In this study, we explored one of the basic approaches, i.e., data resampling, where we oversampled images from the landslide class (i.e., sampling with replacement) to create a balanced training set.

\smallskip\noindent\textbf{Other training details.} We ran all our experiments on Nvidia Tesla P100 GPUs with 16GB memory using PyTorch library.\footnote{\url{https://pytorch.org/}} We adjusted the batch size according to each CNN architecture in order to maximize GPU memory utilization. We used a fixed step size of 50 epochs in the learning rate scheduler of the SGD optimizer and a fixed patience of 50 epochs in the \emph{`ReduceLROnPlateau'} scheduler of the Adam optimizer, both with a factor of 0.1. All of the models were initialized using the weights pretrained on ImageNet~\cite{ILSVRC15} and trained for a total of 200 epochs. Consequently, we trained a total of 560 CNN models in our quest for the best model configuration.

\section{Results}

Due to limited space, Table~\ref{tab:leaderboard} presents results only for the best performing 10 model configurations ranked based on F1-scores obtained on the validation set. The top-performing model configuration (i.e., arch: ResNet50, opt: Adam, lr: $10^{-4}$, wd: $10^{-3}$, no class balancing) achieves an F1-score of 0.805 and an overall accuracy of 0.913, which is deemed plausible performance by the landslide specialists.

\begin{table*}[]
\centering
\begin{tabular}{@{}ccccccccc@{}}
\toprule
\textbf{Optimizer} &
  \textbf{Architecture} &
  \textbf{\begin{tabular}[c]{@{}c@{}}Class\\ Balancing\end{tabular}} &
  \textbf{\begin{tabular}[c]{@{}c@{}}Learning\\ Rate\end{tabular}} &
  \textbf{\begin{tabular}[c]{@{}c@{}}Weight\\ Decay\end{tabular}} &
  \textbf{Accuracy} &
  \textbf{Precision} &
  \textbf{Recall} &
  \textbf{F1} \\ \midrule
Adam & ResNet50     & No & $10^{-4}$ & $10^{-3}$ & 0.913 & 0.834 & 0.779 & 0.805 \\
Adam & ResNet50     & Yes & $10^{-4}$ & $10^{-5}$ & 0.912 & 0.868 & 0.731 & 0.794 \\
SGD  & ResNet50     & Yes & $10^{-2}$ & $10^{-4}$ & 0.907 & 0.816 & 0.771 & 0.793 \\
SGD  & EfficientNet & No & $10^{-2}$ & $10^{-4}$ & 0.904 & 0.793 & 0.790 & 0.791 \\
SGD  & ResNet50     & No & $10^{-3}$ & $10^{-3}$ & 0.906 & 0.821 & 0.760 & 0.789 \\
Adam & ResNet50     & No & $10^{-4}$ & $10^{-4}$ & 0.906 & 0.826 & 0.753 & 0.788 \\
Adam & ResNet50     & Yes & $10^{-4}$ & $10^{-2}$ & 0.907 & 0.835 & 0.745 & 0.788 \\
SGD  & ResNet101    & No & $10^{-3}$ & $10^{-2}$ & 0.904 & 0.806 & 0.768 & 0.786 \\
SGD  & DenseNet     & No & $10^{-2}$ & $10^{-4}$ & 0.905 & 0.819 & 0.753 & 0.785 \\
Adam & EfficientNet & No & $10^{-3}$ & $10^{-4}$ & 0.905 & 0.819 & 0.753 & 0.785 \\
\bottomrule
\end{tabular}%
\caption{Top-performing 10 configurations based on F1-score on the validation set.}
\label{tab:leaderboard}
\end{table*}

Nevertheless, we investigate the full table of results for interesting patterns and identify the following insights:
\begin{itemize}
    \item When everything but the optimizer is kept fixed, the models trained with the Adam optimizer outperforms the models trained with the SGD optimizer (175 vs.\ 104).
    \item Despite the fact that top-performing model is trained without a class balancing strategy, the overall trend indicates that, while everything else is the same, the models trained with class balancing yield better performance than those trained without class balancing (178 vs.\ 95).
    \item ResNet50 architecture tops the rankings among all CNN architectures by achieving the best average ranking as well as the highest mean F1-score according to Table~\ref{tab:architecture}. However, the overall difference between architectures do not seem to be significant except for InceptionNet which yields a significantly poor performance than others.
    \item The impact of the learning rate on model performance shows opposite trends for different optimizers. As per Table~\ref{tab:learning_rate}, smaller learning rates (e.g., $\{10^{-6},10^{-5},10^{-4}\}$) seem to work better with the Adam optimizer whereas larger learning rates (e.g., $\{10^{-2},10^{-3}\}$) seem to work better with the SGD optimizer.
    \item As expected, the value of the weight decay also impacts the overall performance significantly (in particular, for the Adam optimizer). A large weight decay (e.g., $10^{-2}$) hurts the overall performance which tends to improve as the weight decay takes on smaller values (see Table~\ref{tab:weight_decay}).
\end{itemize}

\begin{table}[]
\centering
\begin{tabular}{@{}lccc@{}}
\toprule
\textbf{Architecture} & \textbf{mean(F1)} & \textbf{std(F1)} & \textbf{Avg. Rank}\\ \midrule
ResNet50 & 0.6308 & 0.2262 & 2.70 \\
DenseNet & 0.6256 & 0.2131 & 3.19 \\
ResNet101 & 0.6259 & 0.2146 & 3.28 \\
VGG16 & 0.6114 & 0.2293 & 3.74 \\
EfficientNet & 0.6029 & 0.2089 & 3.90 \\
ResNet18 & 0.6003 & 0.2176 & 4.48 \\
InceptionNet & 0.4001 & 0.1933 & 6.73 \\
\bottomrule
\end{tabular}
\caption{Performance comparison of CNN architectures}
\label{tab:architecture}
\end{table}

\begin{table}[]
\centering
\begin{tabular}{@{}ccccc@{}}
\toprule
\multicolumn{2}{c}{\textbf{Adam}} & \textbf{Learning Rate} & \multicolumn{2}{c}{\textbf{SGD}} \\ \midrule
(mean) & (std) &  & (mean) & (std) \\
0.6843 & 0.0904 & $10^{-6}$ & 0.2168 & 0.2102 \\
0.7001 & 0.0890 & $10^{-5}$ & 0.4292 & 0.2339 \\
0.7203 & 0.0954 & $10^{-4}$ & 0.6600 & 0.1130 \\
0.6298 & 0.1382 & $10^{-3}$ & 0.7185 & 0.0778 \\
0.3791 & 0.2525 & $10^{-2}$ & 0.7145 & 0.0858 \\
\bottomrule
\end{tabular}
\caption{Effect of the learning rate on overall performance}
\label{tab:learning_rate}
\end{table}

\begin{table}[]
\centering
\begin{tabular}{@{}ccccc@{}}
\toprule
\multicolumn{2}{c}{\textbf{Adam}} & \textbf{Weight Decay} & \multicolumn{2}{c}{\textbf{SGD}} \\ \midrule
(mean) & (std) &  & (mean) & (std) \\
0.6618 & 0.1448 & $10^{-5}$ & 0.5582 & 0.2488 \\
0.6573 & 0.1468 & $10^{-4}$ & 0.5570 & 0.2524 \\
0.6319 & 0.1646 & $10^{-3}$ & 0.5569 & 0.2495 \\
0.5400 & 0.2657 & $10^{-2}$ & 0.5192 & 0.2598 \\
\bottomrule
\end{tabular}
\caption{Effect of the weight decay on overall performance}
\label{tab:weight_decay}
\end{table}

\begin{figure*}
    \centering
    \begin{minipage}{0.67\textwidth}
        \begin{subfigure}[b]{0.5\textwidth}
            \centering
            \includegraphics[width=\textwidth]{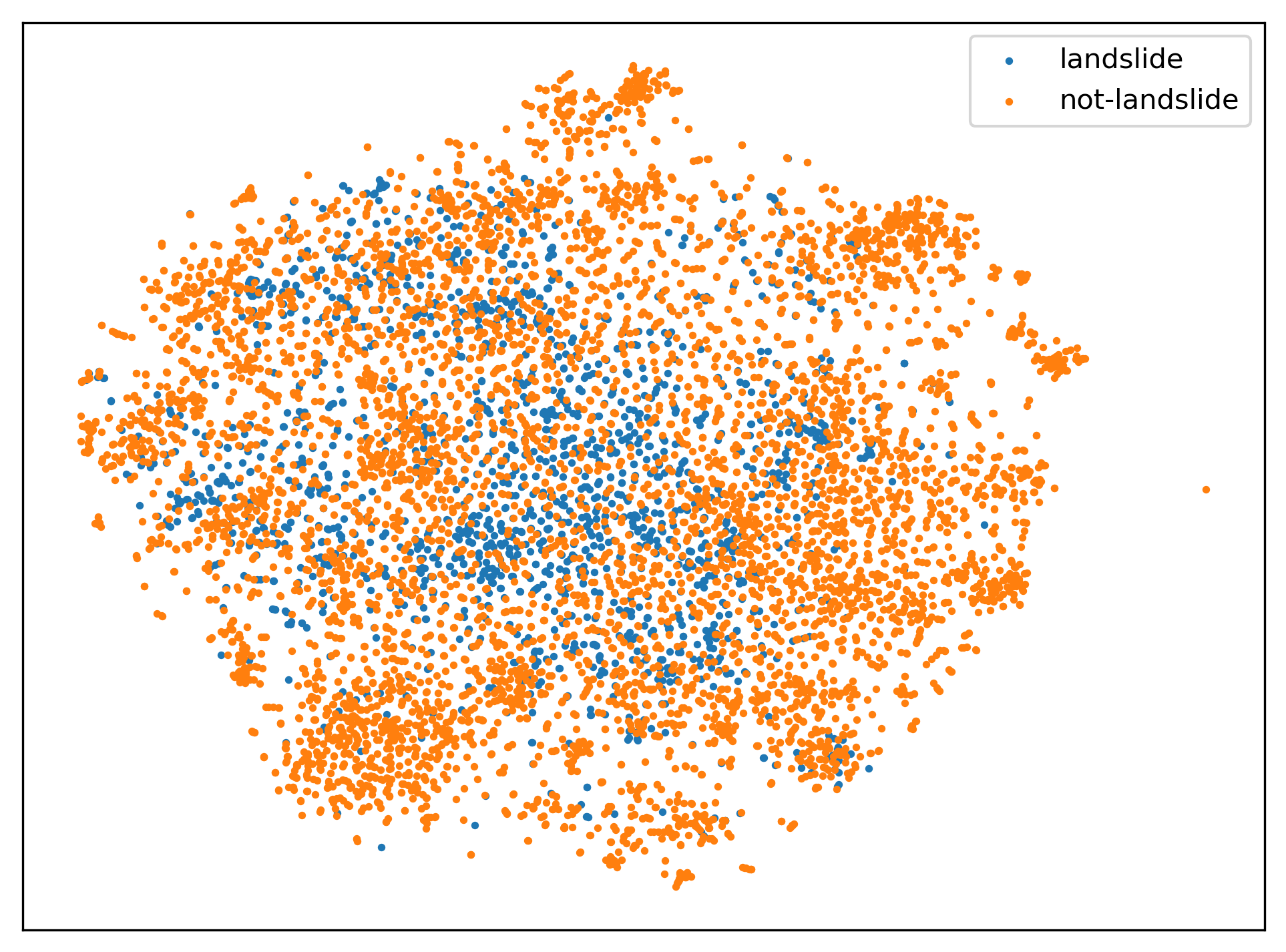}
            \caption[]%
            {{\scriptsize Training set: ResNet50 trained on ImageNet}}
            \label{fig:feat_viz_train_before}
        \end{subfigure}
        \hfill
        \begin{subfigure}[b]{0.5\textwidth}  
            \centering 
            \includegraphics[width=\textwidth]{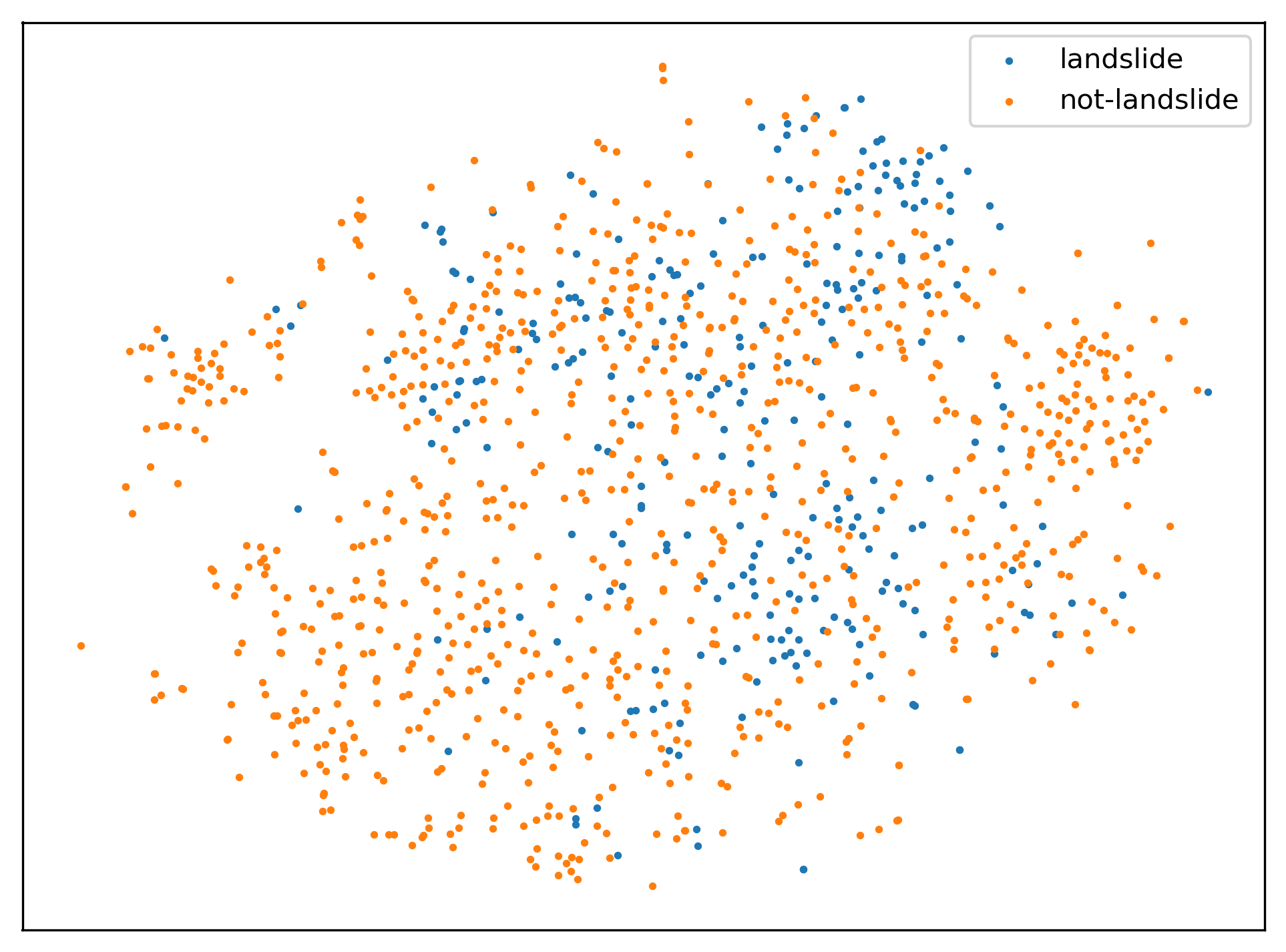}
            \caption[]%
            {{\scriptsize Validation set: ResNet50 trained on ImageNet}}
            \label{fig:feat_viz_val_before}
        \end{subfigure}
        \vskip\baselineskip
        \begin{subfigure}[b]{0.5\textwidth}   
            \centering 
            \includegraphics[width=\textwidth]{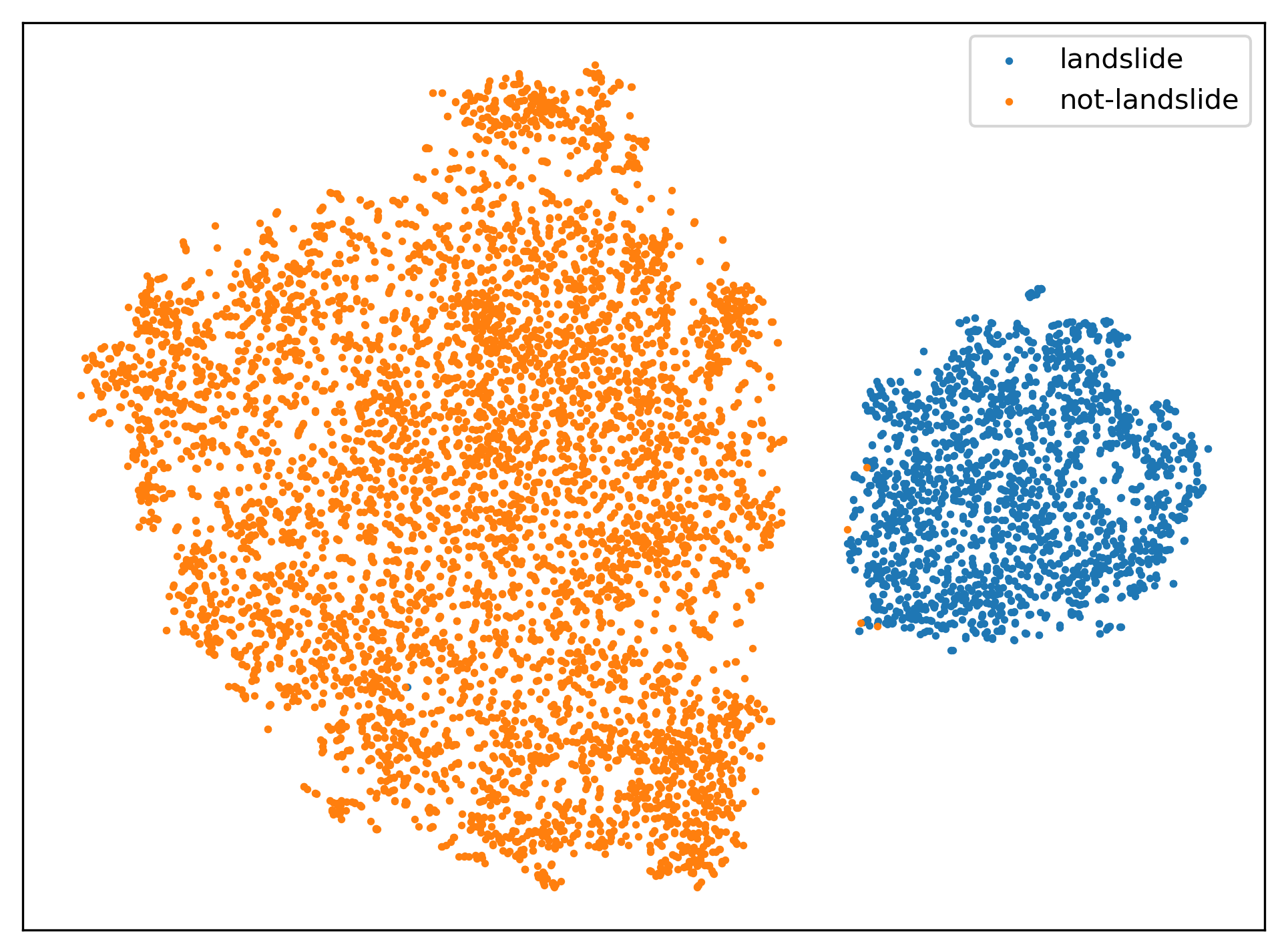}
            \caption[]%
            {{\scriptsize Training set: ResNet50 finetuned on our dataset}}
            \label{fig:feat_viz_train_after}
        \end{subfigure}
        \hfill
        \begin{subfigure}[b]{0.5\textwidth}   
            \centering 
            \includegraphics[width=\textwidth]{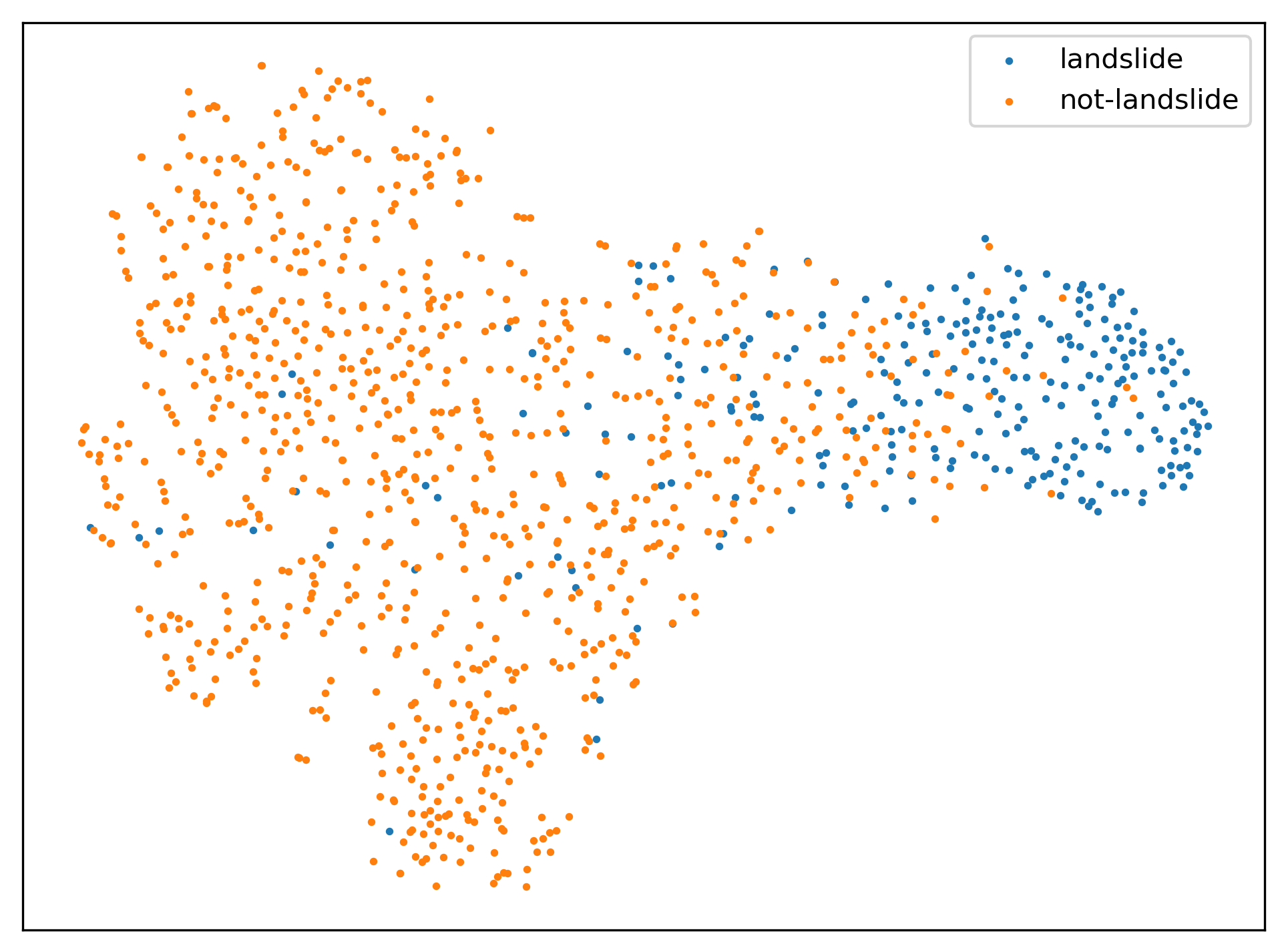}
            \caption[]%
            {{\scriptsize Validation set: ResNet50 finetuned on our dataset}}
            \label{fig:feat_viz_val_after}
        \end{subfigure}
    \end{minipage}%
    \caption[ Visualization of the feature embeddings before/after model finetuning ]
    {Visualization of the feature embeddings before/after model finetuning} 
    \label{fig:feat_viz}
\end{figure*}

To illustrate the success of the transfer learning approach employed in this study, we created t-SNE~\cite{tsne} visualizations of the feature embeddings before and after the training of the best-performing model. As can be seen in Figure~\ref{fig:feat_viz}, the original ResNet50 model pretrained on ImageNet cannot distinguish landslide images from not-landslide images neither in the training set (Figure~\ref{fig:feat_viz_train_before}) nor in the validation set (Figure~\ref{fig:feat_viz_val_before}). However, after finetuning the model on the target landslide dataset, the resulting feature embeddings show almost perfect separation of the classes in the training set (Figure~\ref{fig:feat_viz_train_after}) and a reasonably well separation in the validation set (Figure~\ref{fig:feat_viz_val_after}).

When applied on the held-out test set, the best-performing model achieves an F1-score of 0.701 and an accuracy of 0.870 as opposed to the F1 and accuracy scores of 0.805 and 0.913, respectively, on the validation set (Table~\ref{tab:val_test}). Although the difference in accuracy is relatively small, the difference in F1 is considerably large due to significant drops in precision and recall scores of the model on the test set. This phenomenon can be explained by the more-than-twice increase in the false positive (128 vs.\ 42) and false negative (178 vs.\ 60) predictions of the model on the test set as shown in Table~\ref{tab:conf_mats}.

\begin{figure*}
	\renewcommand{\arraystretch}{0.6} 
	\linespread{0.5}\selectfont\centering
	\resizebox{0.99\linewidth}{!}{%
		\begin{tabular}{p{0.25\textwidth} p{0.25\textwidth} p{0.25\textwidth} p{0.25\textwidth}}
		    \multicolumn{2}{c}{\textbf{TRUE POSITIVES}} & \multicolumn{2}{c}{\textbf{FALSE NEGATIVES}}\\
		    \cmidrule(lr){1-2}\cmidrule(lr){3-4}
			\includegraphics[width=0.12\textwidth]{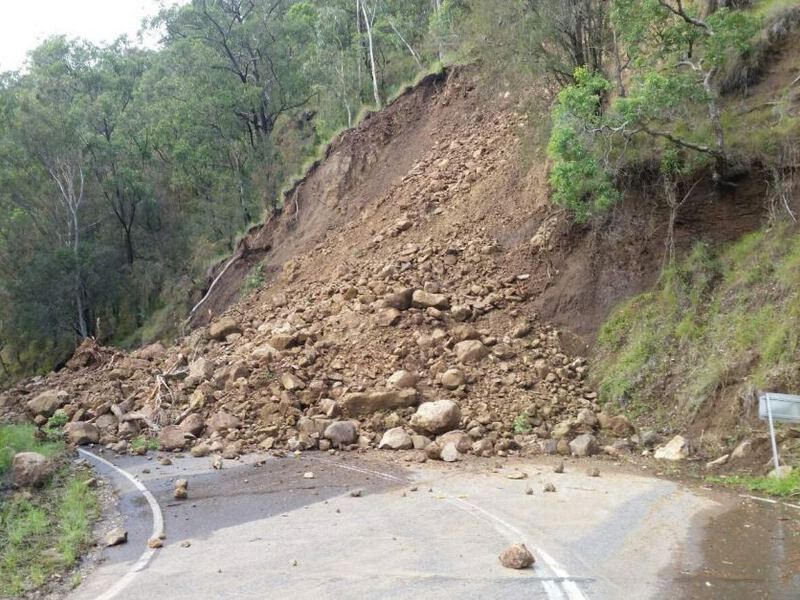}
			\includegraphics[width=0.12\textwidth]{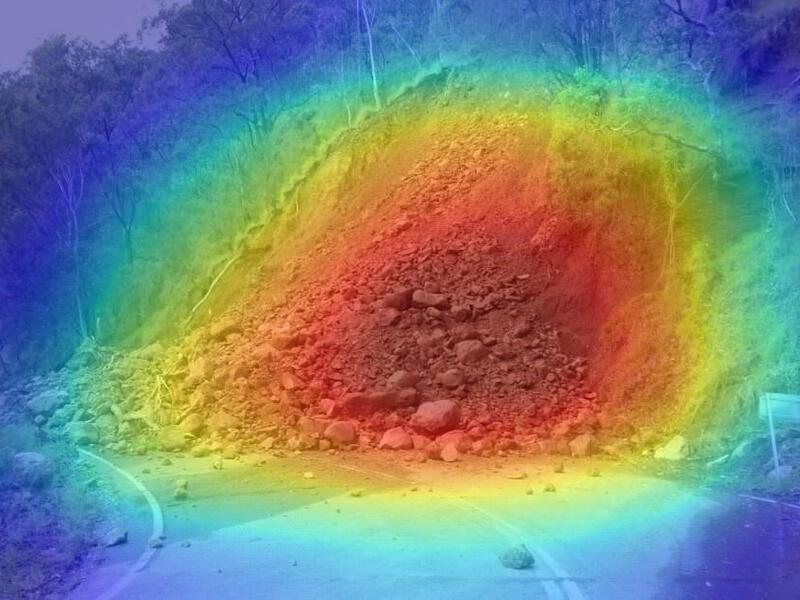}
			&
			\includegraphics[width=0.12\textwidth]{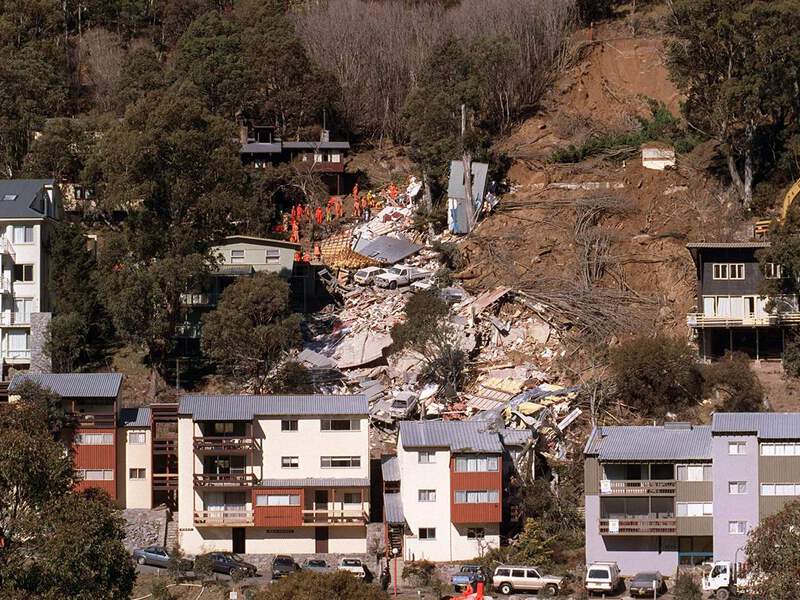}
			\includegraphics[width=0.12\textwidth]{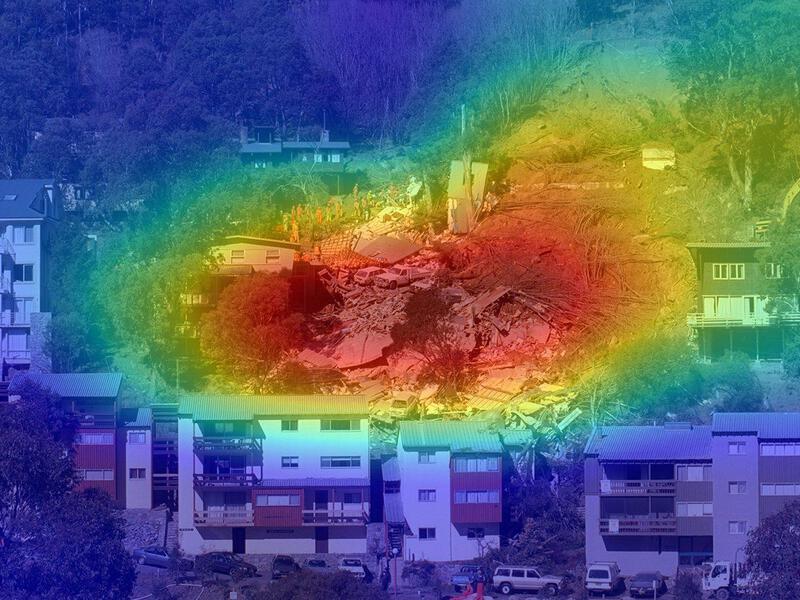}
			&
			\includegraphics[width=0.12\textwidth]{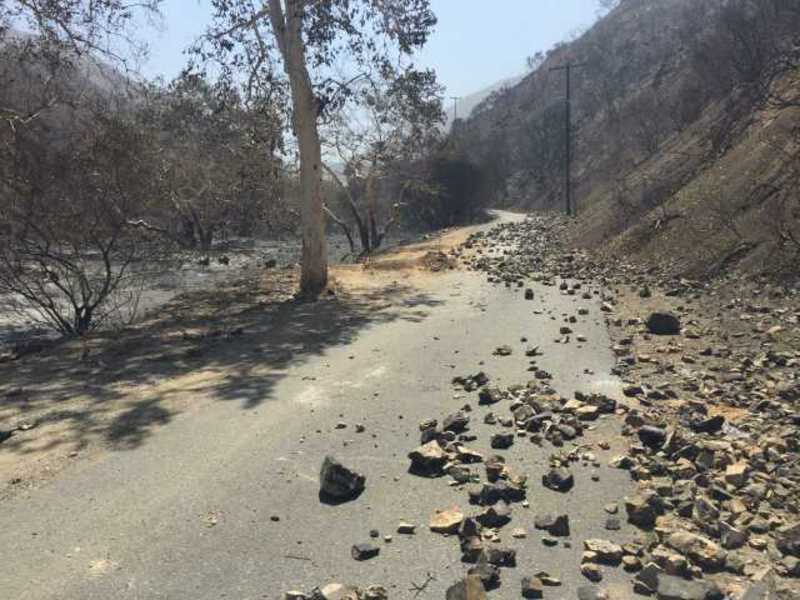}
			\includegraphics[width=0.12\textwidth]{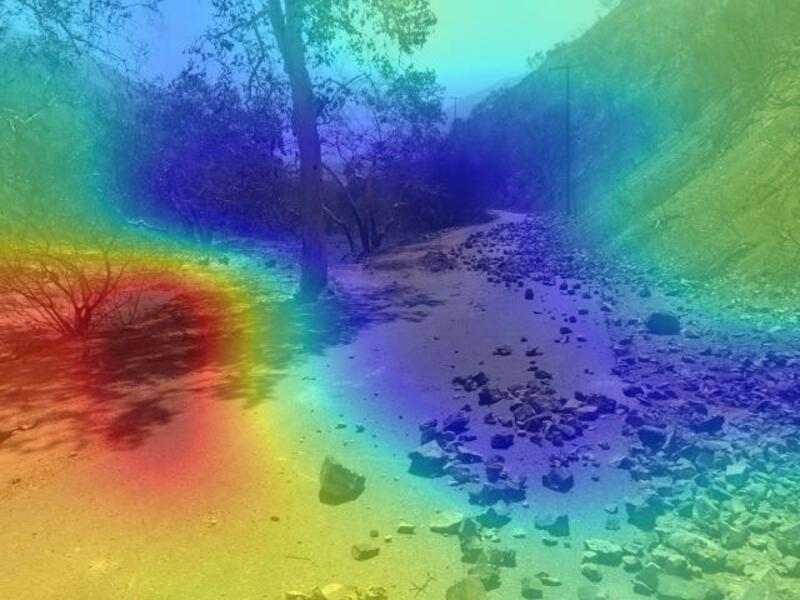}
			&
			\includegraphics[width=0.12\textwidth]{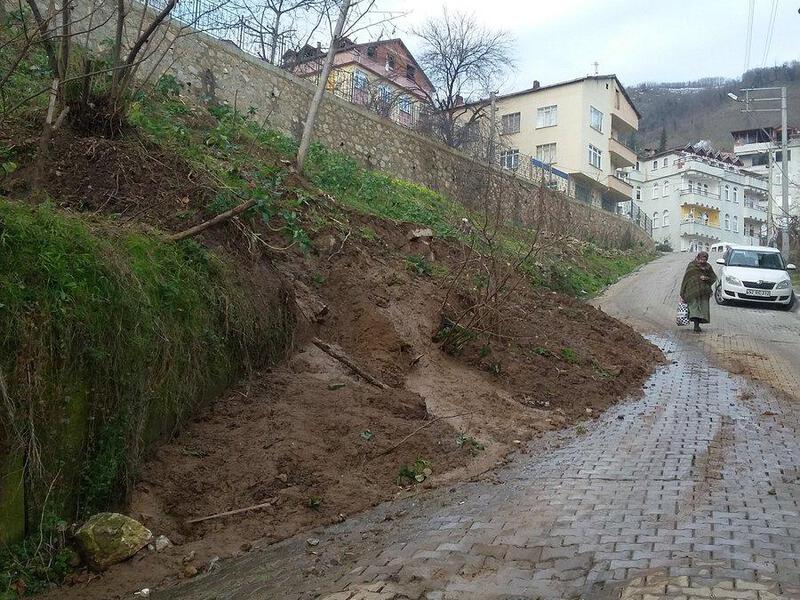}
			\includegraphics[width=0.12\textwidth]{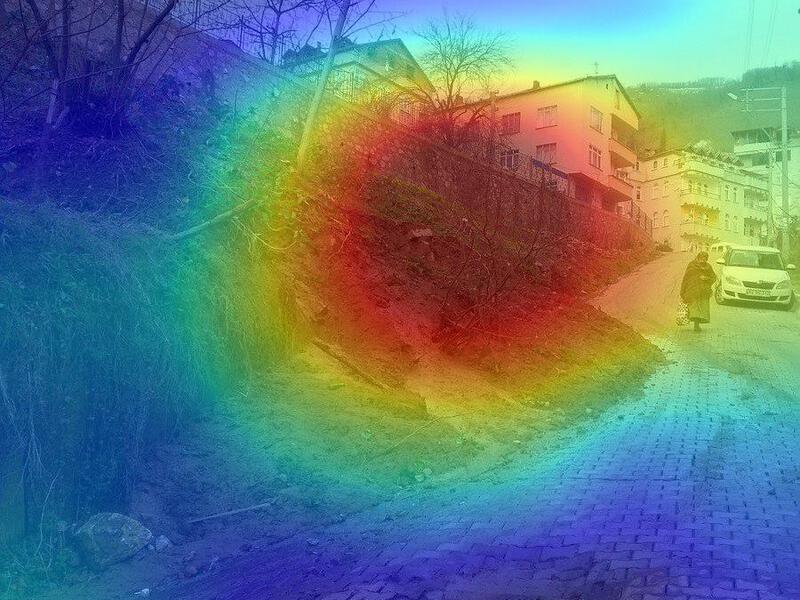}
			\\
			\includegraphics[width=0.12\textwidth]{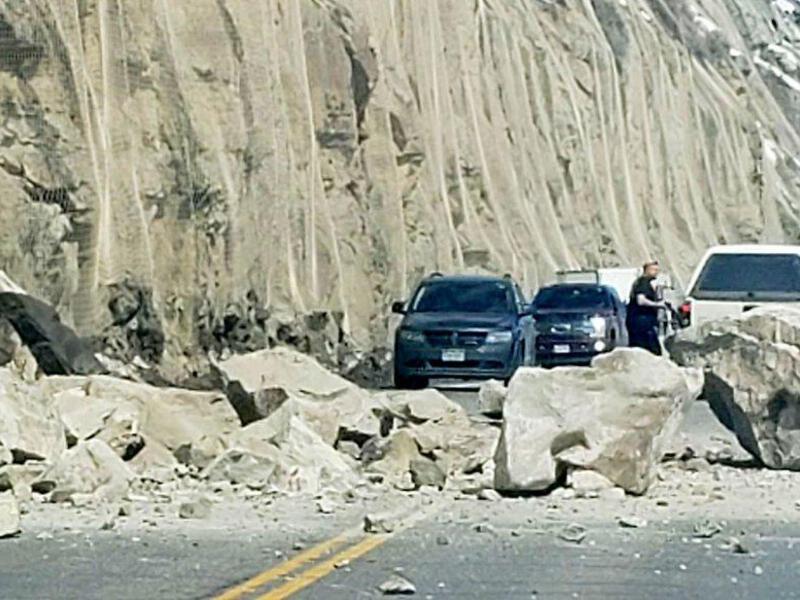}
			\includegraphics[width=0.12\textwidth]{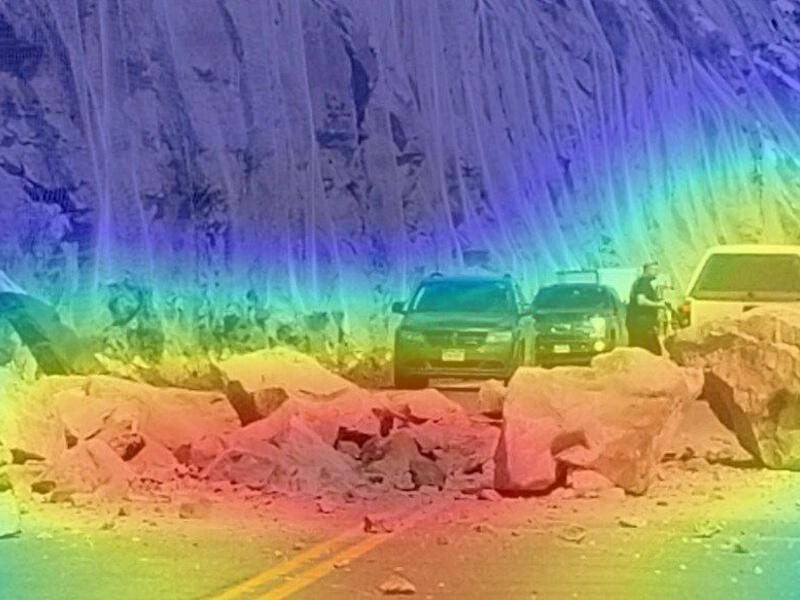}
			&
			\includegraphics[width=0.12\textwidth]{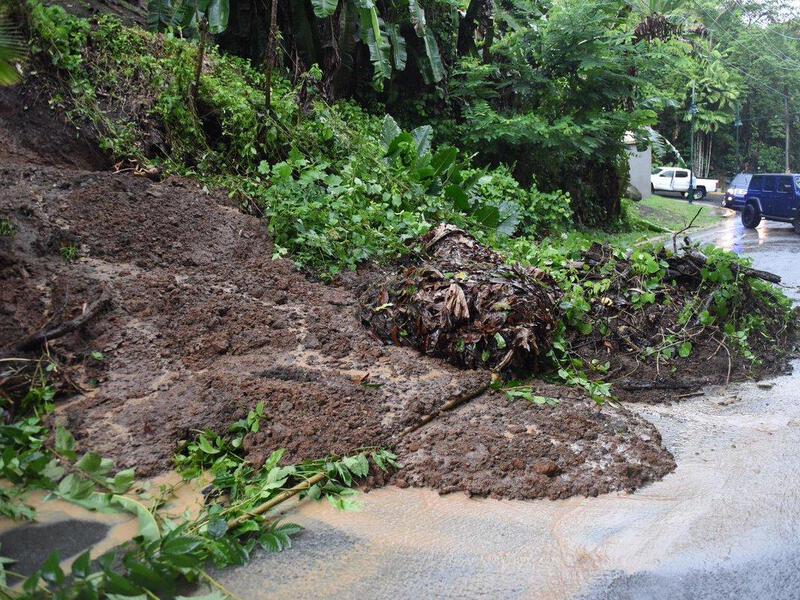}
			\includegraphics[width=0.12\textwidth]{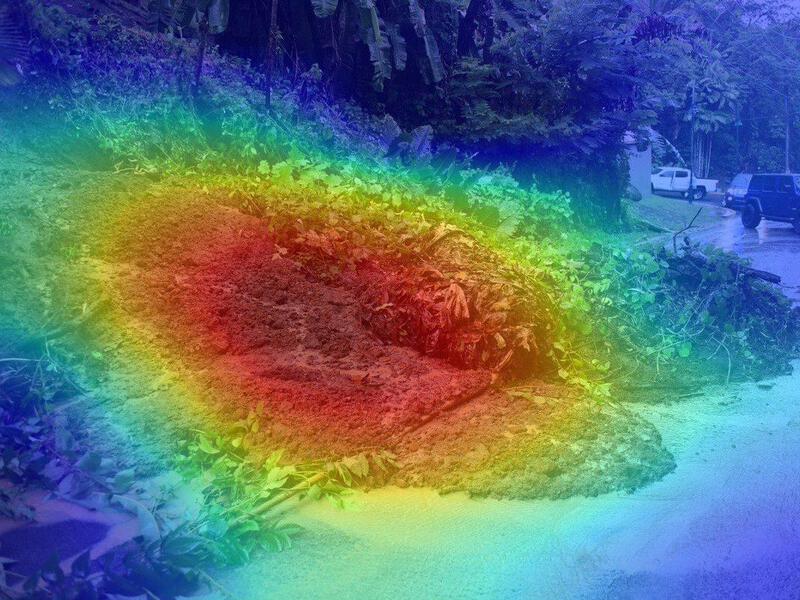}
			&
			\includegraphics[width=0.12\textwidth]{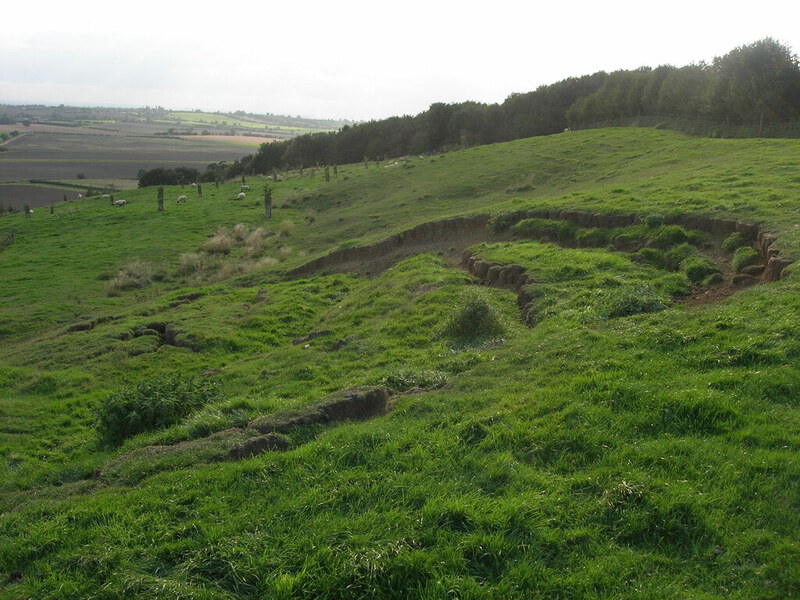}
			\includegraphics[width=0.12\textwidth]{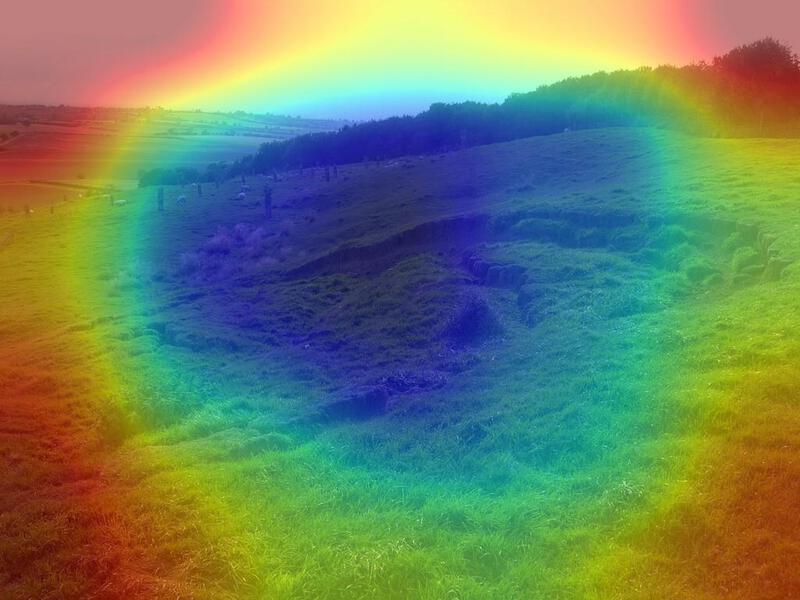}
			&
			\includegraphics[width=0.12\textwidth]{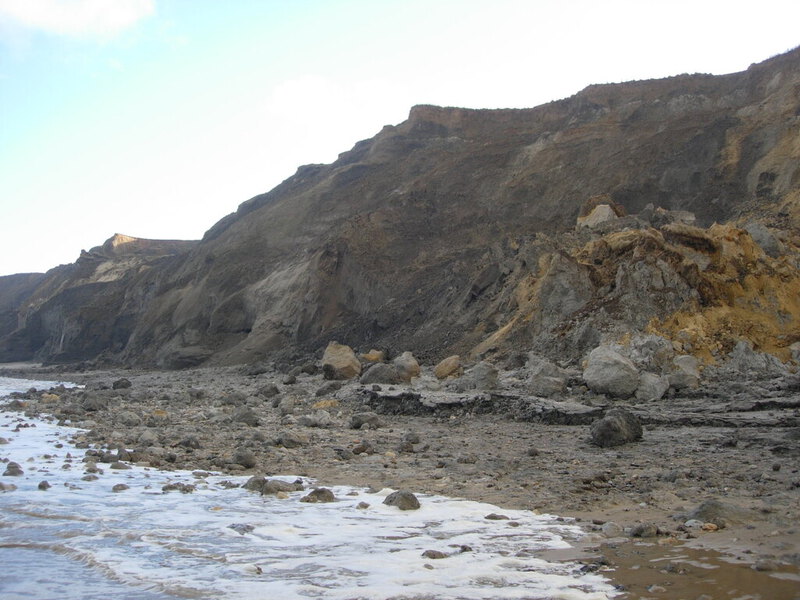}
			\includegraphics[width=0.12\textwidth]{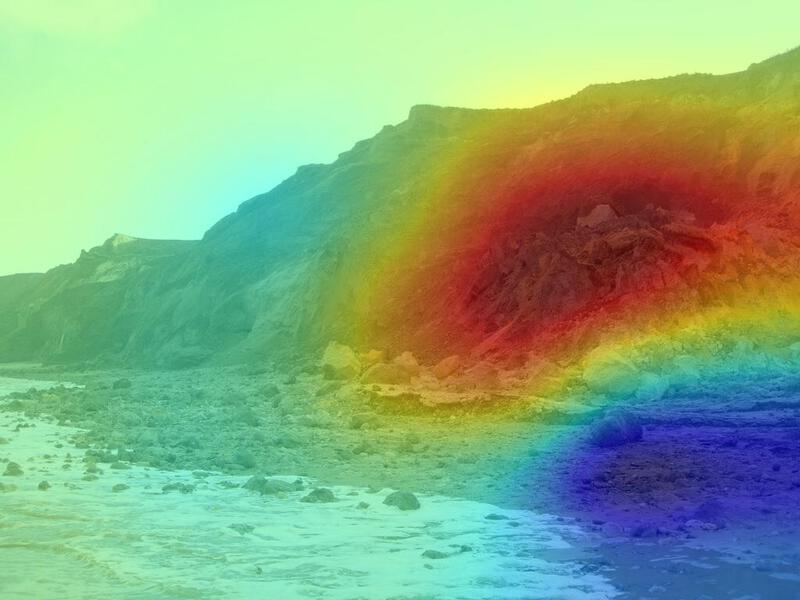}
			\\
			\multicolumn{2}{c}{} & \multicolumn{2}{c}{}\\
			\multicolumn{2}{c}{\textbf{FALSE POSITIVES}} & \multicolumn{2}{c}{\textbf{TRUE NEGATIVES}}\\
		    \cmidrule(lr){1-2}\cmidrule(lr){3-4}
			\includegraphics[width=0.12\textwidth]{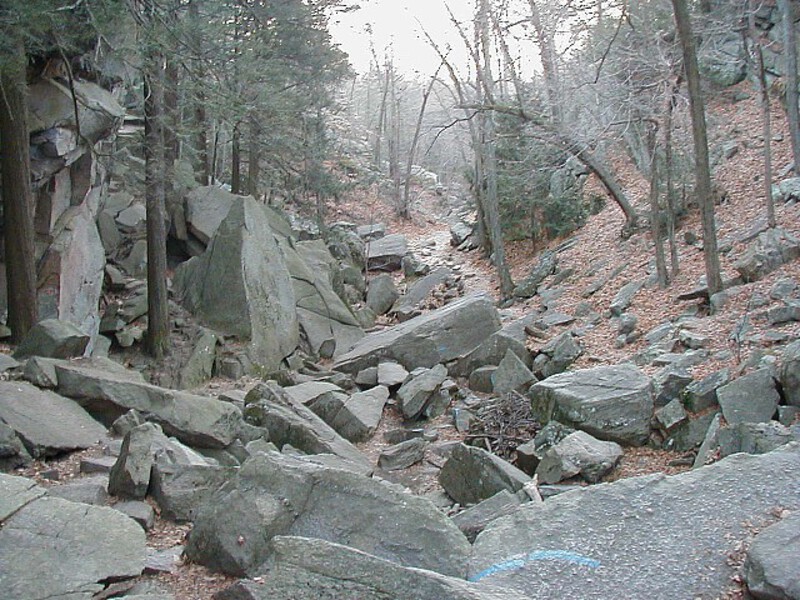}
			\includegraphics[width=0.12\textwidth]{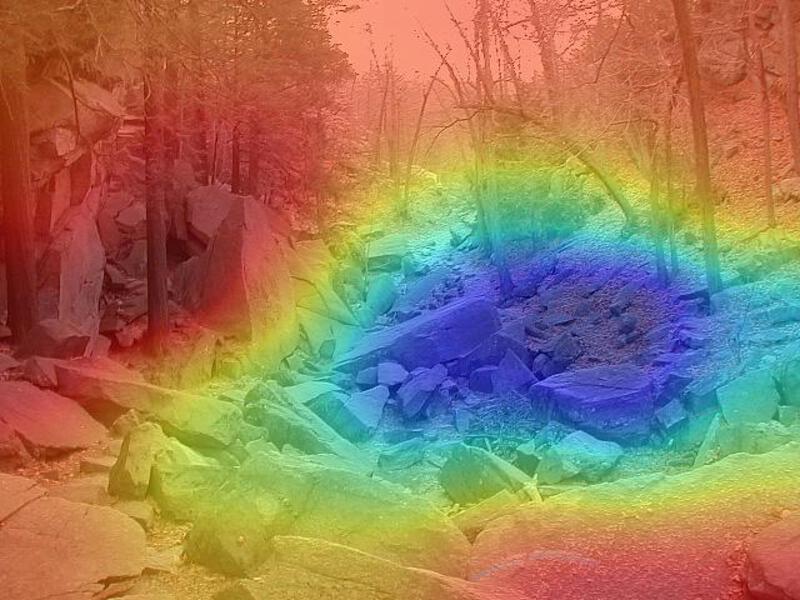}
			&
			\includegraphics[width=0.12\textwidth]{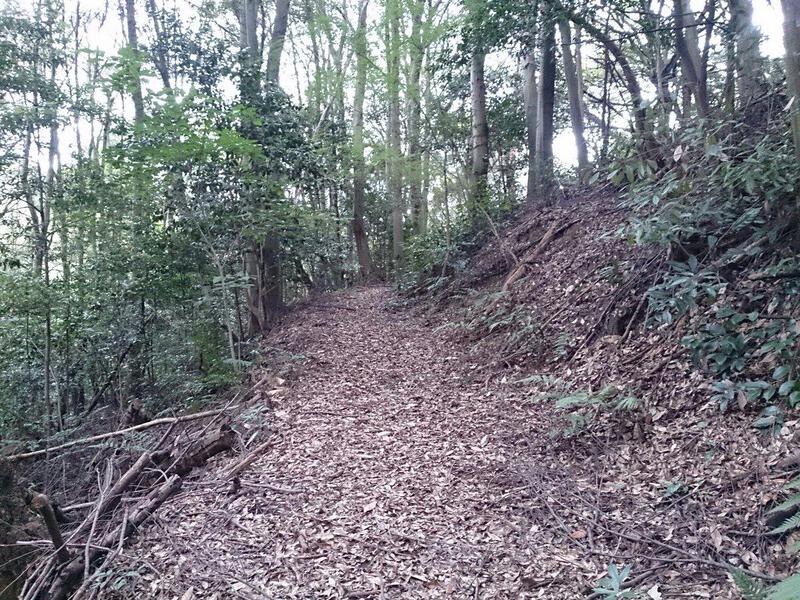}
			\includegraphics[width=0.12\textwidth]{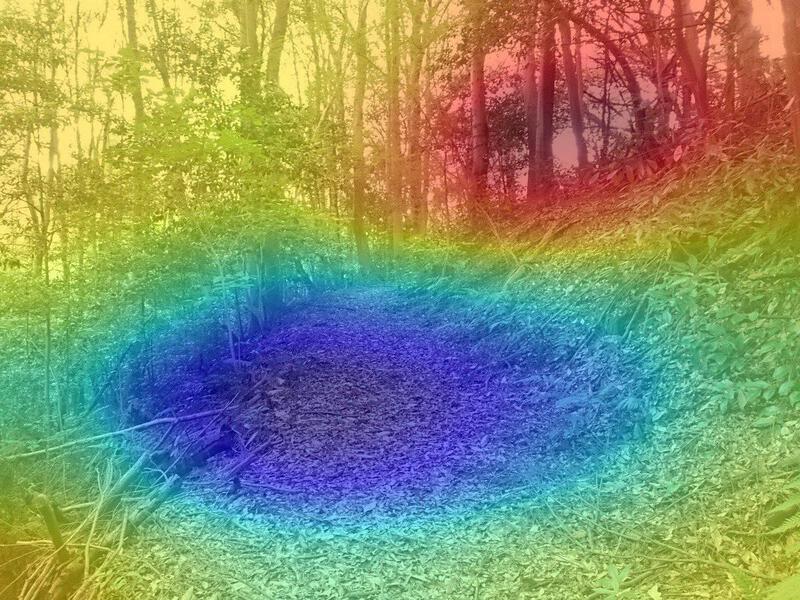}
			&
			\includegraphics[width=0.12\textwidth]{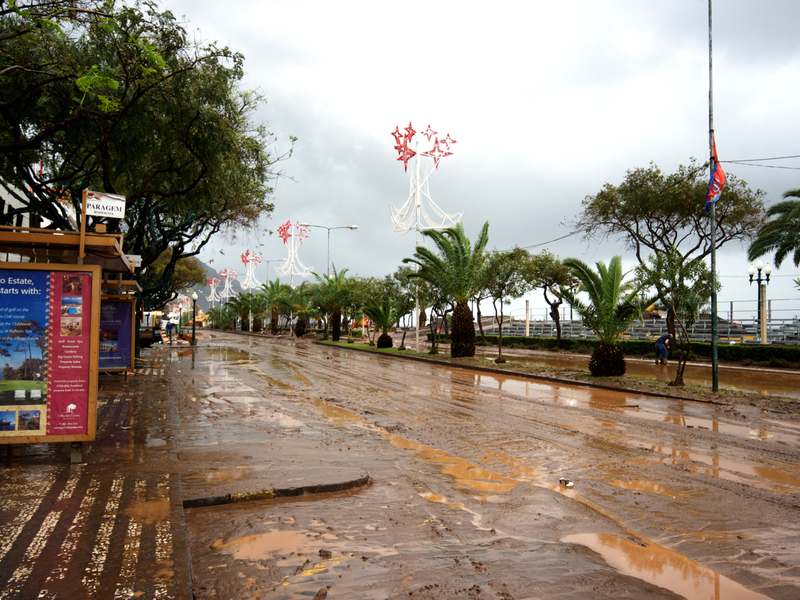}
			\includegraphics[width=0.12\textwidth]{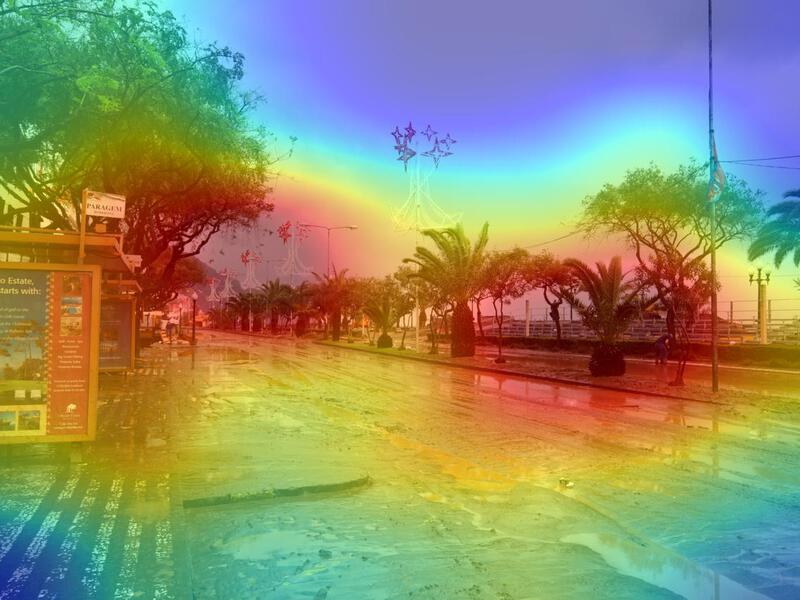}
			&
			\includegraphics[width=0.12\textwidth]{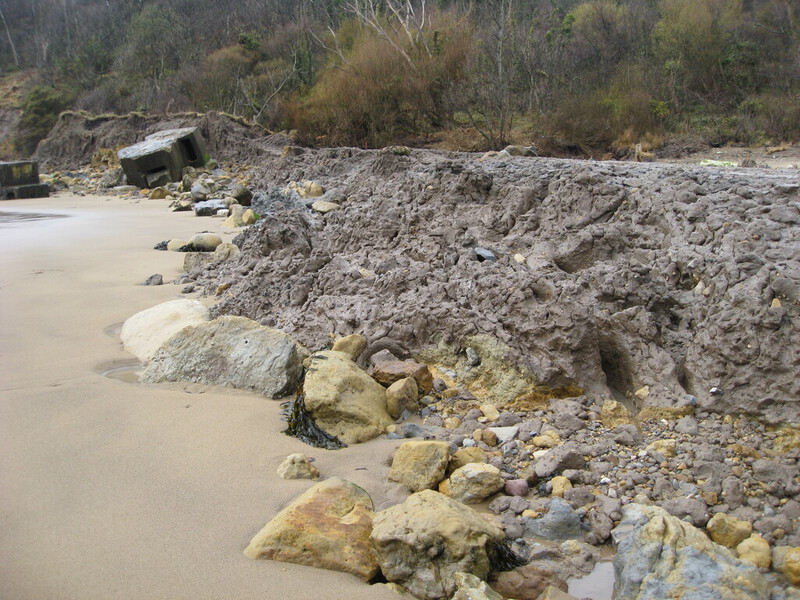}
			\includegraphics[width=0.12\textwidth]{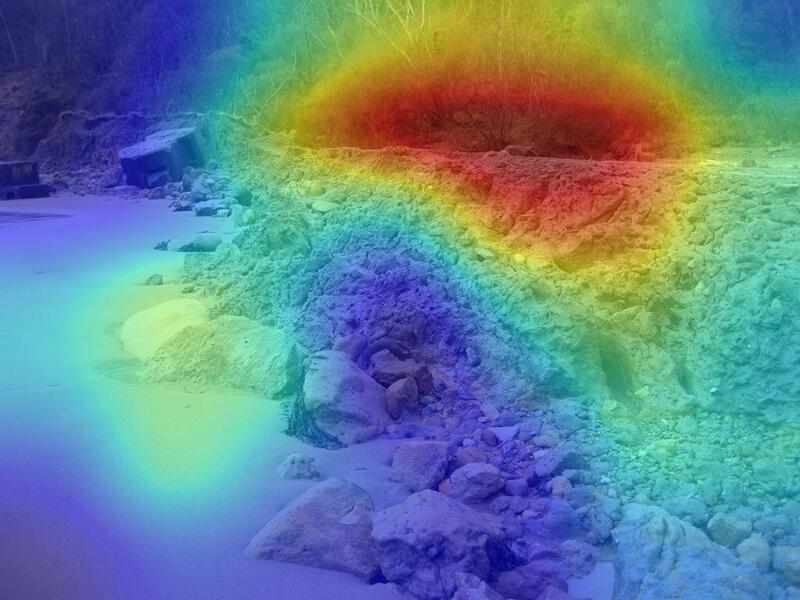}
			\\
			\includegraphics[width=0.12\textwidth]{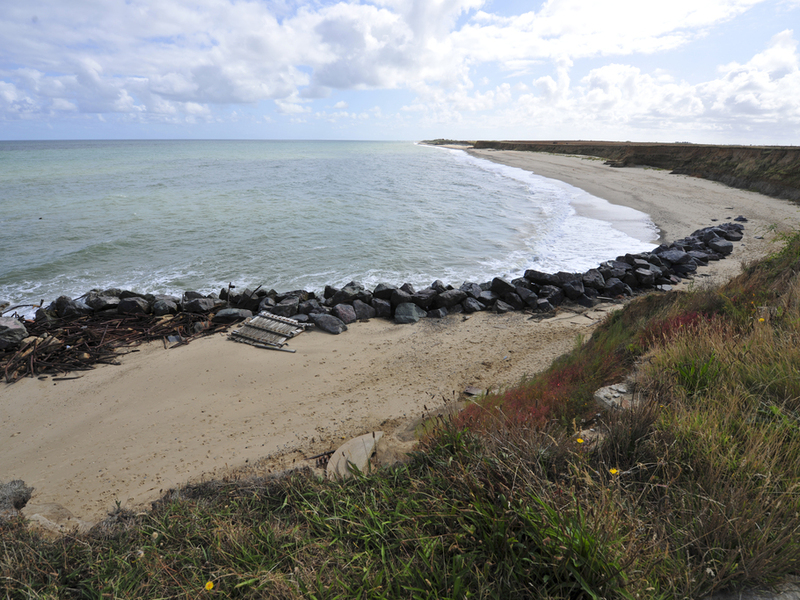}
			\includegraphics[width=0.12\textwidth]{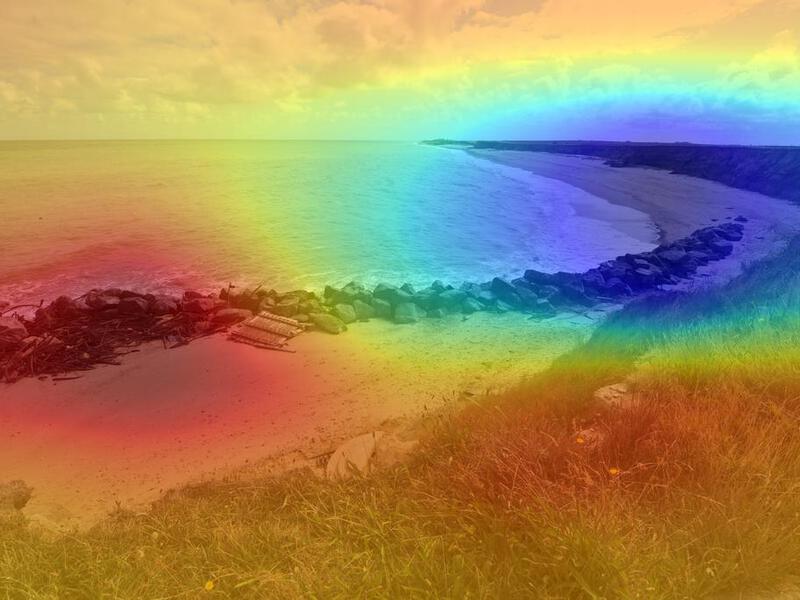}
			&
			\includegraphics[width=0.12\textwidth]{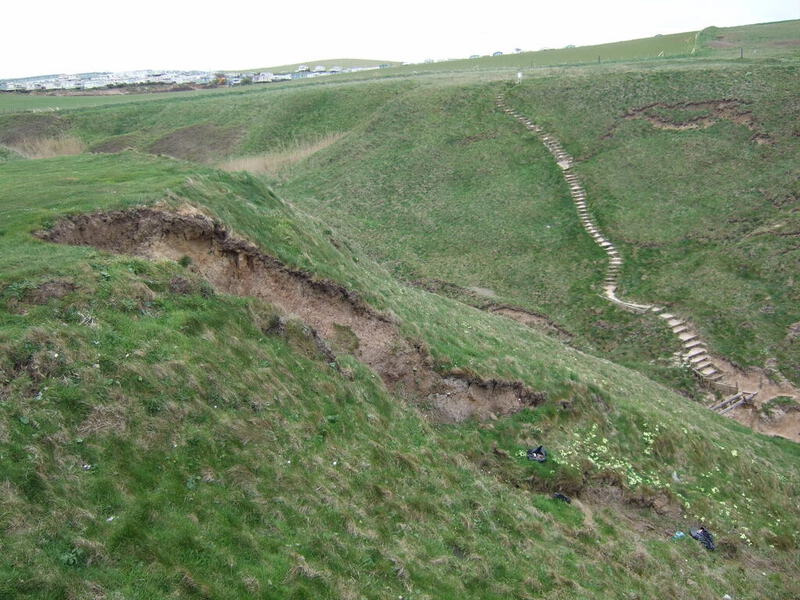}
			\includegraphics[width=0.12\textwidth]{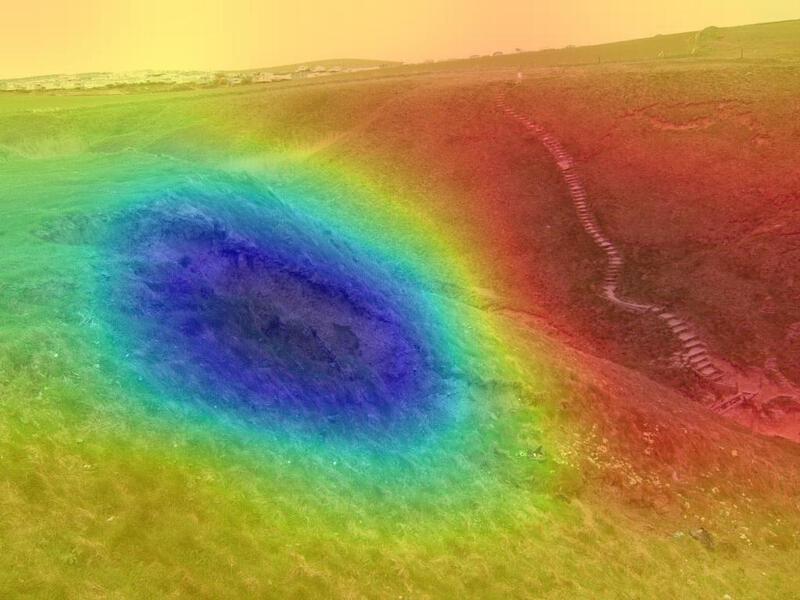}
			&
			\includegraphics[width=0.12\textwidth]{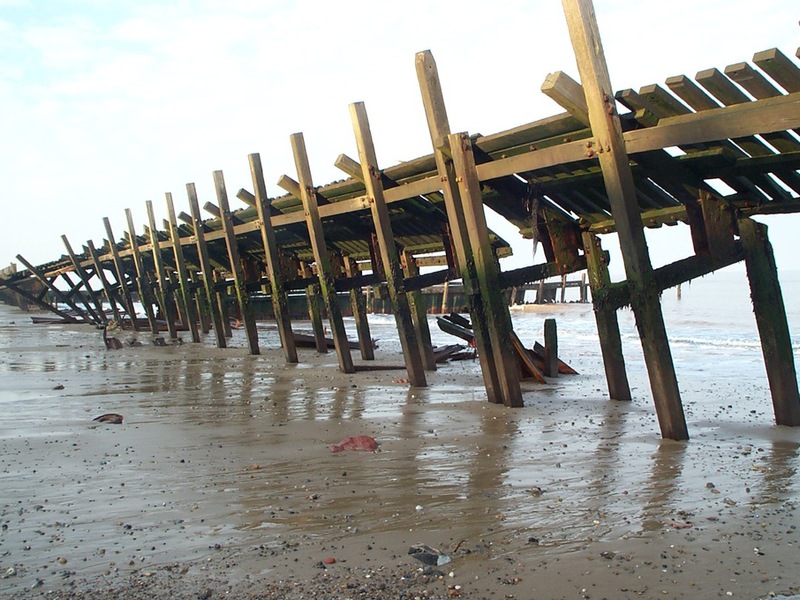}
			\includegraphics[width=0.12\textwidth]{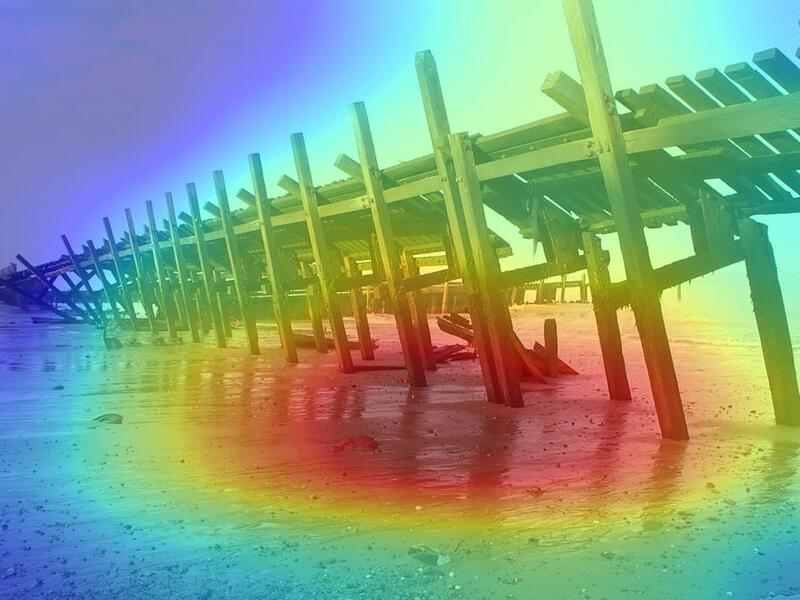}
			&
			\includegraphics[width=0.12\textwidth]{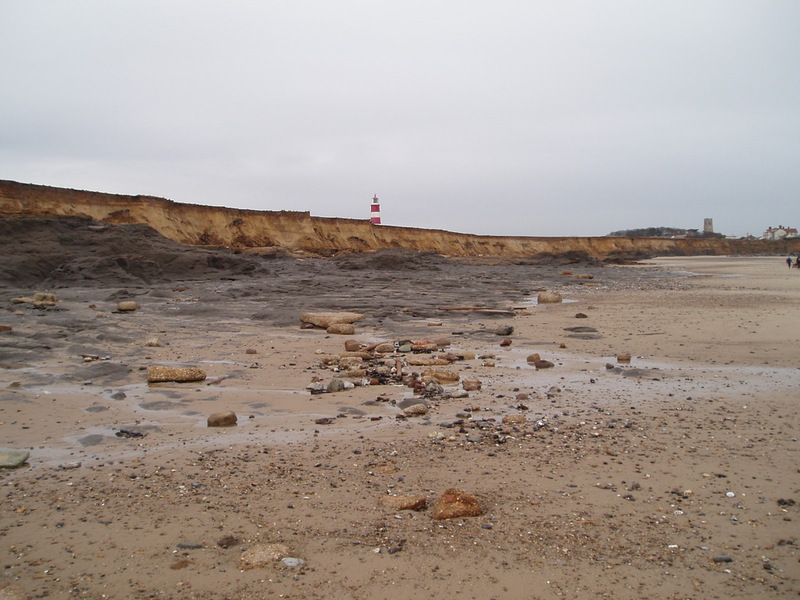}
			\includegraphics[width=0.12\textwidth]{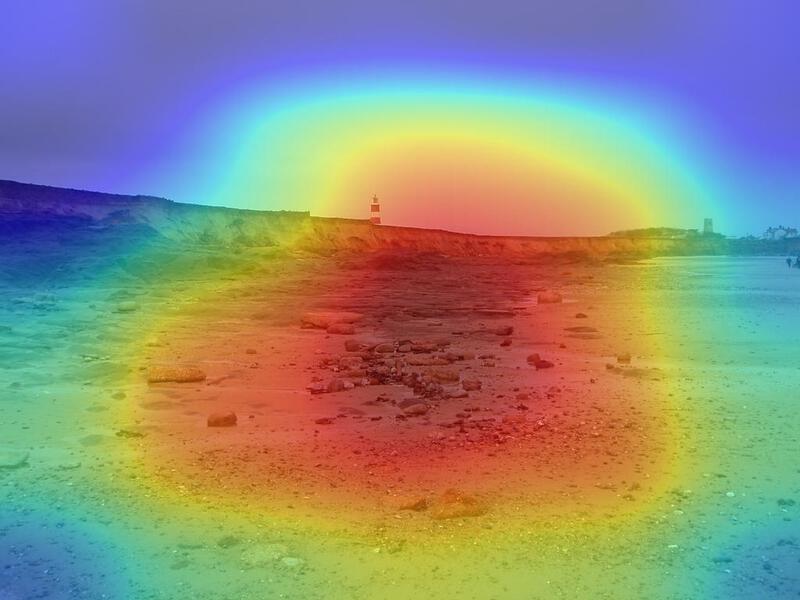}
			\\ 
		\end{tabular}
	}%
	\caption{Class activation map visualizations of the model predictions on the test set.}
	\label{fig:cam_viz}
\end{figure*}

To have a better understanding of the inner workings of the model, we investigated class activation maps~\cite{zhou2016learning}, which highlight the discriminative image regions that the CNN model pays attention to decide whether an image belongs to landslide or not-landslide class. Figure~\ref{fig:cam_viz} demonstrates example visualizations for all four cases, i.e., true positives, true negatives, false positives, and false negatives. The visualizations for the true positive predictions indicate that the model successfully localizes the landslide regions (e.g., rockfalls, earthslip, etc.) in the images. Similarly for the true negative predictions, the model focuses on areas that do not show any landslide cues, successfully avoiding tricky conditions such as muddy roads, wet surfaces, and natural rocky areas on a beach. However, in both false positive and false negative predictions, we observe that the errors occur mainly because the model fails to localize its attention on a particular region in the image, or is tricked by the image regions that are reminiscent of landslide scenes.

\begin{table}[]
\centering
\begin{tabular}{@{}ccccc@{}}
\toprule
\textbf{Set} & \textbf{Accuracy} & \textbf{Precision} & \textbf{Recall} & \textbf{F1} \\ \midrule
Validation & 0.913 & 0.834 & 0.779 & 0.805 \\
Test & 0.870 & 0.737 & 0.668 & 0.701 \\
\bottomrule
\end{tabular}
\caption{Performance comparison of the best model on validation and test sets.}
\label{tab:val_test}
\end{table}

\begin{table}[]
\centering
\begin{tabular}{@{}llrr@{}}
\toprule
 & & \multicolumn{2}{c}{\textbf{Prediction}} \\
 & \textbf{Ground Truth} & Landslide & Not-landslide \\
\midrule
\multirow{2}{*}{\textbf{V (10\%)}} & Landslide & 211 & 60 \\
 & Not-landslide & 42 & 860 \\
\midrule
\multirow{2}{*}{\textbf{T (20\%)}} & Landslide & 358 & 178 \\
 & Not-landslide & 128 & 1,685 \\
\bottomrule
\end{tabular}
\caption{Confusion matrices for the validation and test sets}
\label{tab:conf_mats}
\end{table}

\section{Path to Deployment}
We envision a system that continuously monitors social media (i.e., Twitter) for general landslide-related content and deploys our landslide classification model to identify and retain the most relevant information. The planned system will follow a design approach similar to the one presented in \cite{alam2017image4act,alam2018processing} without the human-in-the-loop aspect (for now). Specifically, there will be a Tweet Collector module that will collect live tweets from the Twitter Streaming API~\footnote{\url{https://developer.twitter.com/en/docs/tutorials/consuming-streaming-data}} that matches landslide-related keywords and hashtags in multiple languages. This module will be followed by an Image Collector module that will extract image URLs from the tweets (if any) and download images. Then, the Image Classifier module will run the downloaded images through our landslide model to tag each image as landslide or not-landslide. In parallel, the Geolocation Inference module will use tweet metadata to geolocate the images following the approach presented in~\cite{qazi2020geocov19}. Eventually, all the results will be stored in a database by the Persister module, which will then be used by the Visualizer Module to create a dashboard and/or a map representation of the detection results. With this plan, we hope to translate this fruitful collaboration between researchers and practitioners into a solid outcome that can benefit the landslide community as well as the government agencies and humanitarian organizations.

\section{Conclusion}

In this study, we aimed to develop a model that can automatically detect landslides in social media image streams. For this purpose, we created a large image collection from multiple sources with different characteristics to ensure data diversity. Then, the collected images were assessed by three experts to attain high quality labels with almost substantial inter-annotator agreement. At the heart of this study lied an extensive search for the optimal landslide model configuration with various CNN architectures, network optimizers, learning rates, weight decays, and class balancing strategies. We provided several insights about the impact of each optimization dimension on the overall performance. The best-performing model achieved high performance in terms of accuracy and F1 scores, which can be deemed sufficient for the purpose. Furthermore, presented error analyses pointed at potential improvements for future work. Finally, we described a road map to deploy the proposed landslide model in an online, real-time system.

\bibliography{main}

\end{document}